\documentclass{article}

    \PassOptionsToPackage{numbers, compress}{natbib}
\usepackage[preprint]{neurips_2025}

\usepackage[utf8]{inputenc} % allow utf-8 input
\usepackage[T1]{fontenc}    % use 8-bit T1 fonts
\usepackage{hyperref}       % hyperlinks
\usepackage{url}            % simple URL typesetting
\usepackage{booktabs}       % professional-quality tables
\usepackage{graphicx}       % includegraphics
\usepackage{caption}        % captionof
\usepackage{amsmath}        % math environments and \boldsymbol
\usepackage{amssymb}        % additional math symbols
\usepackage{amsfonts}       % blackboard math symbols
\usepackage{nicefrac}       % compact symbols for 1/2, etc.
\usepackage{microtype}      % microtypography
\usepackage{xcolor}         % colors
\usepackage{wrapfig}        % inline figures and tables
\usepackage{subcaption}     % subfigures
\usepackage{enumitem}       % list customization
\usepackage{multirow}       % multi-row table cells
\usepackage{colortbl}       % colored table rows
\usepackage{algorithm}
\usepackage{algorithmic}
\usepackage{listings}
\title{RiT: Vanilla Diffusion Transformers Suffice in Representation Space}

\author{
  Le Zhang$^{1}$\thanks{Corresponding author: \texttt{le.zhang@mila.quebec}} \quad Ning Mang$^{2}$ \quad Aishwarya Agrawal$^{1,3}$ \\
  $^{1}$Mila -- Qu\'ebec AI Institute, UdeM \quad $^{2}$Utrecht University \quad $^{3}$Canada CIFAR AI Chair
}

\begin{document}

\maketitle

\begin{abstract}
  Flow matching with $x$-prediction---regressing the clean data point rather than the ambient velocity---is known to exploit low-dimensional manifold structure effectively in pixel space \cite{li2025back}. We ask whether a pretrained representation space, while containing a low-dimensional data manifold of comparable intrinsic dimensionality, offers a distribution more favorable for flow-matching learning. Comparing pixel, SD-VAE, and DINOv2 features along four geometric axes, we find that pixel and DINOv2 share nearly identical intrinsic dimensionalities (both $\hat{d}\!\approx\!33$) yet DINOv2 exhibits $7.3\times$ higher effective rank, $35\times$ better covariance conditioning, $11.5\times$ lower excess kurtosis, and $1.7\times$ lower on-manifold interpolation error; SD-VAE latents are consistently intermediate, indicating that the advantage stems from representation-learning objectives rather than mere compression. These statistical properties render the flow-matching regression well-conditioned and remove the need for the specialized prediction heads or Riemannian transport used by prior DINOv2 diffusion methods. We propose the \emph{Representation Image Transformer} (RiT): a vanilla Diffusion Transformer trained by $x$-prediction on frozen DINOv2 features, augmented only by a dimension-aware noise schedule and joint \texttt{[CLS]}-patch modeling. On ImageNet $256{\times}256$, RiT attains FID 1.45 without guidance and 1.14 with classifier-free guidance, outperforming DiT$^\text{DH}$-XL with $19\%$ fewer parameters (676M vs.\ 839M). The resulting ODE is efficiently solvable at coarse discretizations: with classifier-free guidance, $5$ Heun steps already reach FID 2.0 and $10$ steps reach 1.25, without distillation or consistency training. Code at \url{https://github.com/lezhang7/RiT}.
\end{abstract}

\section{Introduction}

Flow matching \cite{liu2022flow, esser2024scaling} learns a velocity field that transports Gaussian noise to data along linear paths. When data concentrates near a low-dimensional manifold, $x$-prediction---parameterizing the network to output the clean data point $\hat{\mathbf{z}}_0$ rather than the ambient-space velocity---places the regression target on that manifold, as demonstrated by JiT \cite{li2025back} in pixel space. A natural question is whether a pretrained representation space, while containing a data manifold of comparable intrinsic dimensionality, offers a distribution more favorable for learning the flow-matching velocity field.

Comparing pixel, SD-VAE \cite{rombach2022high}, and DINOv2 \cite{oquab2024dinov2} features along four geometric axes, we find that pixel and DINOv2 share nearly identical intrinsic dimensionalities (both $\hat{d}\!\approx\!33$) yet embed this manifold differently relative to $\mathcal{N}(\mathbf{0}, \mathbf{I})$. The pixel manifold is anisotropic, has strongly non-Gaussian per-coordinate marginals, and admits linear chords that traverse low-density regions. DINOv2 features exhibit near-isotropic variance, near-Gaussian per-coordinate marginals \cite{wang2020understanding}, and approximately on-manifold linear interpolants. These are marginal properties, not joint ones: DINOv2 features still concentrate on a $\hat{d}\!\approx\!33$-dimensional manifold, but each coordinate's transport toward $\mathcal{N}(\mathbf{0},\mathbf{I})$ is short and well-conditioned.

Section~\ref{sec:manifold} quantifies these gaps: DINOv2 attains $7.3\times$ higher effective rank, $35\times$ better covariance conditioning, $11.5\times$ lower excess kurtosis, and $1.7\times$ lower on-manifold interpolation error than pixels. SD-VAE latents fall consistently between the two, indicating that the advantage arises from representation-learning objectives rather than compression alone. These distributional advantages coexist with a DINOv2-specific pathology at off-manifold intermediate states: per-token LayerNorm pins $\|\mathbf{z}\|\!\approx\!\sqrt{d}$, so linear flow-matching paths $\mathbf{z}_t = t\mathbf{z}_0 + (1{-}t)\boldsymbol{\epsilon}$ traverse ambient regions the encoder never outputs, and the $v$-target at such $\mathbf{z}_t$ acquires a large radial component.

The prevailing response to this radial ambiguity has been architectural. RAE \cite{zheng2025diffusion} handles it with a specialized wide prediction head (DDT \cite{ddt}) atop $v$-prediction, alongside a ViT decoder that maps DINOv2 features back to pixels. Concurrent work \cite{kumar2025rjf} calls this phenomenon \emph{geometric interference} and replaces the Euclidean transport with Riemannian Flow Matching on the norm-concentration sphere. Both modifications add complexity to either the architecture or the transport path.

We take a target-side alternative: $x$-prediction. Under this parameterization, the network regresses $\hat{\mathbf{z}}_0$, which lies on the data manifold by construction, so the radial ambiguity is resolved at the network's output (which targets $\mathbf{z}_0$ on the manifold) rather than at its input (where $\mathbf{z}_t$ remains off-manifold). The reparameterization itself is not new \cite{li2025back}; its effectiveness here stems from the combination with DINOv2's isotropic per-coordinate variance and near-Gaussian marginals, which render the denoising regression $\mathbf{z}_t \to \mathbf{z}_0$ well-conditioned enough for a vanilla DiT. We instantiate this combination as the \textbf{Representation Image Transformer (RiT)} (Section~\ref{sec:method}): a vanilla Diffusion Transformer trained by $x$-prediction flow matching in representation space, augmented by a dimension-aware noise schedule and joint \texttt{[CLS]}-patch modeling. As DiT operates on the SD-VAE latent space, RiT operates on a representation space provided by a frozen encoder--decoder; we use RAE's \citep{zheng2025diffusion} frozen DINOv2 encoder and ViT decoder. RiT thus models the high-dimensional DINOv2 feature distribution directly, without adapting the encoder for generation. On ImageNet $256^2$, RiT attains FID 1.45 without guidance and 1.14 with classifier-free guidance, outperforming DiT$^\text{DH}$-XL with $19\%$ fewer parameters. The resulting ODE converges in few Heun steps, yielding 5-step FID 2.0 and 10-step FID 1.25 (guided) without distillation or consistency training (Section~\ref{sec:few_step}).

\section{The Geometry of Representation Spaces for Flow Matching}\label{sec:manifold}

\begin{figure}[htb]
  \centering
  \begin{minipage}[t]{0.32\textwidth}
    \centering
    \includegraphics[width=\textwidth]{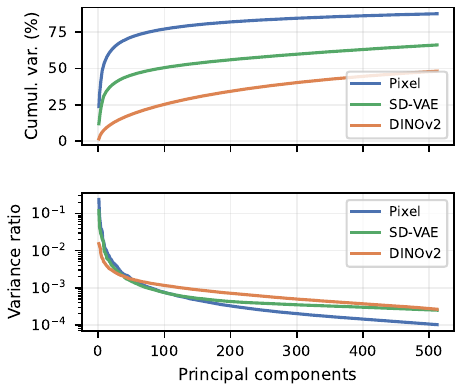}
    \vspace{-6mm}
    \caption*{(a) PCA spectrum}
  \end{minipage}
  \hfill
  \begin{minipage}[t]{0.32\textwidth}
    \centering
    \includegraphics[width=\textwidth]{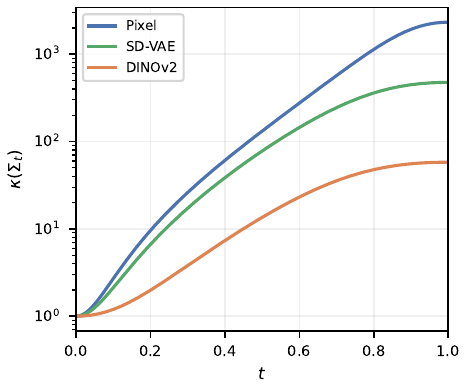}
    \vspace{-6mm}
    \caption*{(b) Optimization conditioning}
  \end{minipage}
  \hfill
  \begin{minipage}[t]{0.32\textwidth}
    \centering
    \includegraphics[width=\textwidth]{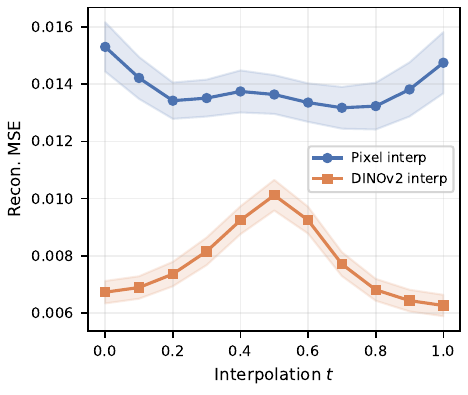}
    \vspace{-6mm}
    \caption*{(c) On-manifold interpolation}
  \end{minipage}
  \caption{\textbf{Manifold analysis across Pixel, SD-VAE, and DINOv2.} (a) PCA spectrum: cumulative variance (top) and per-component variance on log scale (bottom); flatter decay indicates more uniform spread. (b) Condition number $\kappa(\Sigma_t)$ along the transport path; DINOv2 stays $35\times$ better conditioned than Pixel at $t{=}0.9$. (c) Interpolation reconstruction MSE; Pixel stays off-manifold while DINOv2 remains close throughout.}
  \label{fig:manifold_analysis}
  \vspace{-5pt}
\end{figure}

The manifold hypothesis---that data concentrates on a low-dimensional surface---holds regardless of representation. What differs across representations is how \emph{favorably} this manifold is positioned relative to $\mathcal{N}(\mathbf{0}, \mathbf{I})$, and thus whether transport paths are short and the ODE is efficiently solvable in few steps. We characterize each representation along four complementary axes---\textbf{intrinsic dimensionality} (the manifold's true degrees of freedom), \textbf{effective rank} (how uniformly variance is spread across directions), \textbf{marginal Gaussianity} (per-coordinate similarity to $\mathcal{N}(0,1)$), and \textbf{on-manifold linear interpolation} (whether linear chords stay near data)---each predicting a distinct mechanism that makes flow matching easier or harder. We quantify these on three spaces over 10{,}000 ImageNet images: (i)~raw pixels ($3 {\times} 256 {\times} 256$, $D{=}196{,}608$), (ii)~DINOv2-Base features ($768 {\times} 16 {\times} 16$, $D{=}196{,}608$), and (iii)~SD-VAE latents ($4 {\times} 32 {\times} 32$, $D{=}4{,}096$)---the pretrained VAE used as the latent space of latent diffusion models \cite{rombach2022high}. Pixels and DINOv2 share the same ambient dimensionality, enabling direct geometric comparison; the inclusion of SD-VAE isolates the effect of representation-learning training (exemplified by DINOv2's SSL) from generic compression.

\noindent\textbf{Intrinsic dimensionality} is the manifold's true degrees of freedom---the number of independent directions needed to describe the data after stripping away ambient redundancy. Two spaces with comparable intrinsic dimensionality face manifolds of comparable underlying complexity; any difference in flow-matching learning difficulty must therefore arise from \emph{how} the manifold is positioned rather than from its size. We use the TwoNN estimator \citep{facco2017estimating}, which recovers $d$ by maximum likelihood from the ratio of second- to first-nearest-neighbor distances under local uniformity (Appendix~\ref{app:manifold}). Bootstrapping over 10 independent subsamples of 5{,}000 points gives $\hat{d} = 33.6 \pm 1.3$ for pixels and $\hat{d} = 32.6 \pm 0.8$ for DINOv2---nearly identical, with the 1-dimension gap well within the combined estimator standard deviation. Both spaces therefore share essentially the same underlying manifold dimensionality; DINOv2's advantage, by elimination, lies in \emph{how} that manifold is embedded relative to the noise.

\begin{figure}[t]
\centering
\begin{minipage}[c]{0.45\textwidth}
  \centering
  \small
  \resizebox{0.8\textwidth}{!}{
  \begin{tabular}{@{}lccc@{}}
    \toprule
    Metric & Pixel & SD-VAE & DINOv2 \\
    \midrule
    Mean $|\kappa|$ & 0.923 & 0.467 & \textbf{0.114} \\
    Median $|\kappa|$ & 0.958 & 0.228 & \textbf{0.083} \\
    $|\kappa| < 0.5$ & 0.0\% & 74.2\% & \textbf{98.7\%} \\
    $|\kappa| < 1.0$ & 70.6\% & 86.7\% & \textbf{99.7\%} \\
    \bottomrule
  \end{tabular}
  }
  \captionof{table}{\textbf{Marginal Gaussianity.} Excess kurtosis across three representation spaces.}
  \label{tab:kurtosis}
  \vspace{6pt}
  \includegraphics[width=\textwidth]{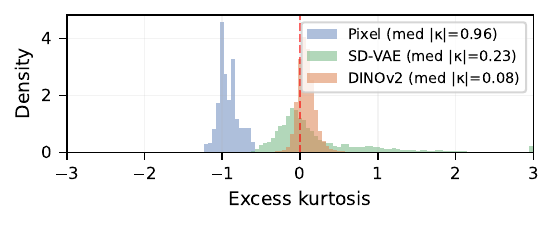}
  \captionof{figure}{\textbf{Kurtosis distribution.} DINOv2 marginals concentrate tightly around $\kappa{=}0$ (Gaussian); SD-VAE is intermediate; pixels deviate strongly.}
  \label{fig:kurtosis}
\end{minipage}
\hfill
\begin{minipage}[c]{0.52\textwidth}
  \centering
  \includegraphics[width=\textwidth]{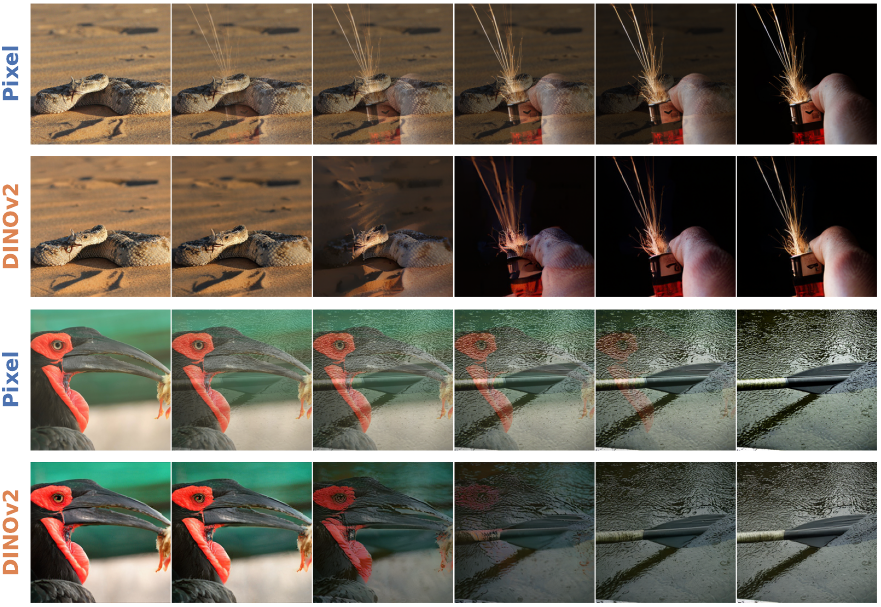}
  \captionof{figure}{\textbf{Cross-class interpolation.} Top row of each pair: pixel-space blending $\mathbf{x}_t {=} (1{-}t)\mathbf{x}_a {+} t\mathbf{x}_b$ (ghosting artifacts). Bottom row: interpolation in DINOv2 representation space $\mathbf{z}_t {=} (1{-}t)\mathbf{z}_a {+} t\mathbf{z}_b$, then decoded back to pixels via the RAE decoder (smooth semantic transitions).}
  \label{fig:interpolation}
\end{minipage}
\vspace{-10pt}
\end{figure}

\noindent\textbf{Effective rank} quantifies how uniformly variance is distributed across principal directions. It equals $1$ when all variance concentrates in a single direction (a thin needle in $\mathbb{R}^D$) and equals the ambient dimension when variance spreads perfectly evenly (an isotropic ball). Since the flow-matching source $\mathcal{N}(\mathbf{0},\mathbf{I})$ is itself isotropic, higher effective rank on the data side translates to shorter, more uniform transport paths from noise to data. Concretely, $\text{erank} = \exp\bigl(-\sum_i \hat\lambda_i \log\hat\lambda_i\bigr)$ with $\hat\lambda_i = \lambda_i/\sum_j \lambda_j$ the normalized PCA eigenvalues \citep{roy2007effective}. Figure~\ref{fig:manifold_analysis}(a) plots per-component variance (log scale) and its cumulative: pixel's first 50 components capture ${\sim}60\%$ of total variance versus ${\sim}25\%$ for DINOv2. The effective ranks are $45$, $98$, and $327$ for pixels, SD-VAE, and DINOv2 respectively---a $7.3\times$ gap between pixels and DINOv2. DINOv2's per-token LayerNorm further fixes $\|\mathbf{z}\|^2 = d$ by construction, so this high effective rank concentrates features near an approximately isotropic shell of radius $\sqrt{d}$ containing the $\hat d\!\approx\!33$-dim data manifold \citep{wang2020understanding,kumar2025rjf}.

\noindent\textbf{Optimization conditioning} reflects whether different variance directions can be learned in parallel during training: a well-conditioned regression converges along all directions at comparable rates, while a poorly-conditioned one over-fits high-variance directions while starving low-variance ones. Flow matching at time $t$ is implicitly such a regression, with effective covariance interpolating between $\mathbf{I}$ at $t{=}0$ (pure noise) and the data covariance $\mathbf{H}$ at $t{=}1$ (clean data); ill-conditioned $\mathbf{H}$ therefore propagates into late-schedule training. Concretely, under a local Gaussian approximation $p(\mathbf{z}_0)\!\approx\!\mathcal{N}(\boldsymbol{\mu},\mathbf{H})$, \citet{ahamed2025preconditioned} show the regression covariance is $\Sigma_t = (1{-}t)^2 \mathbf{I} + t^2 \mathbf{H}$; we use the standard condition number $\kappa(\Sigma_t) = \lambda_{\max}/\lambda_{\min}$ as the diagnostic. Figure~\ref{fig:manifold_analysis}(b) plots $\kappa(\Sigma_t)$ across $t \in [0,1]$: both spaces start at $\kappa{=}1$ near $t{=}0$ (where $\Sigma_t\to\mathbf{I}$) and grow monotonically toward $\kappa(\mathbf{H})$ as $t{\to}1$. At $t{=}0.9$ (representative of late-schedule fine-grained data-fitting), pixel space reaches $\kappa \approx 2{,}000$ while DINOv2 stays at $\kappa \approx 56$---a $35\times$ gap, enabling all variance components to be learned at comparable rates. The same distributional proximity to $\mathcal{N}(\mathbf{0},\mathbf{I})$ also tightens the posterior $p(\mathbf{z}_0\mid\mathbf{z}_t)$, shrinking the irreducible variance $\mathbb{E}\|\mathbf{v}-\mathbb{E}[\mathbf{v}\mid\mathbf{z}_t]\|^2$ of the per-pair velocity target---a distinct mechanism contributing to the faster convergence in \S\ref{sec:experiments}.

\noindent\textbf{Marginal Gaussianity} measures how close each \emph{individual} coordinate's 1D distribution is to a Gaussian. The source $\mathcal{N}(\mathbf{0},\mathbf{I})$ is Gaussian along every axis, so closer-to-Gaussian per-coordinate marginals on the data side keep each dimension's transport from noise to data short and well-behaved. We use the per-dimension excess kurtosis $\kappa_j = \mathbb{E}[(z_j - \mu_j)^4] / \sigma_j^4 - 3$, which is $0$ for a Gaussian, positive for heavier-than-Gaussian (outlier-prone) tails, and negative for lighter tails. As shown in Table~\ref{tab:kurtosis} and Figure~\ref{fig:kurtosis}, DINOv2 dimensions are markedly more Gaussian: 98.7\% satisfy $|\kappa_j| < 0.5$ (vs.\ 74.2\% for SD-VAE and 0\% for pixels), with median $|\kappa_j|$ $11.5\times$ lower than pixels and $2.7\times$ lower than SD-VAE. This captures marginal behavior only; the interpolation experiment below probes the joint geometry.

\noindent\textbf{On-manifold linear interpolation.} The previous three axes summarize variance per-direction; the final axis probes the \emph{joint} geometry. Flow matching transports samples along straight paths $\mathbf{z}_t = t\mathbf{z}_0 + (1{-}t)\boldsymbol{\epsilon}$, so if linear chords between data points themselves wander off the manifold, intermediate $\mathbf{z}_t$ states will too, leaving the velocity target poorly defined. Cross-class image interpolation makes this concrete: pixel interpolation produces ghosting artifacts characteristic of paths crossing low-density voids, while DINOv2 interpolation yields smooth semantic transitions (Figure~\ref{fig:interpolation}). We quantify this via a round-trip reconstruction error (full procedure in Appendix~\ref{app:manifold}): each intermediate frame---whether obtained by pixel blending or by linear interpolation in DINOv2 space followed by RAE decoding---is passed through the \emph{same} DINOv2 encoder--RAE-decoder pipeline \cite{zheng2025diffusion}, and the MSE versus the input measures off-manifold distance. Because both conditions traverse the identical pipeline, the encoder--decoder reconstruction bias is shared; the remaining gap isolates whether the frame lies on or off the image manifold. Pixel frames incur $1.7\times$ higher error than DINOv2 frames ($0.0136$ vs.\ $0.0080$); Figure~\ref{fig:manifold_analysis}(c) shows DINOv2 remains close throughout while pixel stays uniformly off-manifold.

\noindent\textbf{Summary.} Pixel and DINOv2 share nearly identical intrinsic dimensionalities (both $\hat{d}\!\approx\!33$) yet DINOv2 is far better suited to flow-matching learning: $7.3\times$ higher effective rank, $35\times$ better covariance conditioning, $11.5\times$ lower excess kurtosis, and $1.7\times$ lower on-manifold interpolation error; SD-VAE is consistently intermediate, indicating the advantage arises from representation-learning objectives rather than compression alone. These properties predict that DDT heads, Riemannian transports, and wider backbones are not required for competitive performance---a prediction we validate in Sections~\ref{sec:method}--\ref{sec:experiments} with a vanilla DiT and $x$-prediction.

\section{RiT: A Vanilla DiT for Representation-Space Diffusion}\label{sec:method}

\begin{wrapfigure}{r}{0.35\textwidth}
  \centering
  \vspace{-18pt}
  \includegraphics[width=0.35\textwidth]{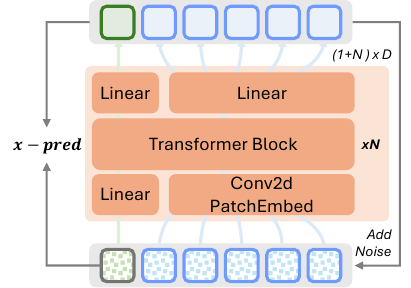}
  \caption{\textbf{RiT Arch.} Frozen RAE encoder/decoder (gray) bracket a vanilla DiT trained by $x$-prediction; [CLS] and patch tokens share self-attention, with separate heads for $\hat{\mathbf z}_0$ and $\hat{\mathbf z}_{\text{cls},0}$.}
  \label{fig:method}
  \vspace{-15pt}
\end{wrapfigure}

Guided by the geometry of Section~\ref{sec:manifold}, we instantiate the \textbf{Representation Image Transformer (RiT)} (Figure~\ref{fig:method}): a vanilla DiT backbone (SwiGLU \cite{shazeer2020glu}, RMSNorm \cite{zhang2019root}, 2D-RoPE \cite{su2024roformer}, QK-normalization \cite{henry2020query}, plus in-context class tokens following JiT \cite{li2025back}), trained with a recipe tailored to DINOv2 features. We reuse RAE's \cite{zheng2025diffusion} frozen DINOv2-with-Registers encoder \cite{oquab2024dinov2} and ViT decoder to move between pixels and features. The encoder yields patch tokens $\mathbf{z} \in \mathbb{R}^{d \times h \times w}$ and a \texttt{[CLS]} token $\mathbf{z}_\text{cls} \in \mathbb{R}^{d}$; both are projected, concatenated, and jointly attended, with separate linear heads predicting $\hat{\mathbf{z}}_0$ and $\hat{\mathbf{z}}_{\text{cls},0}$. Full details in Appendix~\ref{app:architecture}.

\noindent\textbf{Flow matching preliminaries.} Flow matching \cite{liu2022flow, esser2024scaling} learns a velocity field that transports noise $\boldsymbol{\epsilon} \sim \mathcal{N}(\mathbf{0},\mathbf{I})$ to data $\mathbf{z}_0$ along straight paths $\mathbf{z}_t = t\,\mathbf{z}_0 + (1{-}t)\,\boldsymbol{\epsilon}$, with $t \in [0,1]$, $t{=}0$ pure noise and $t{=}1$ clean data; the path's time derivative is $\mathbf{v} = \mathbf{z}_0 - \boldsymbol{\epsilon}$. The standard \emph{$v$-prediction} objective trains a network $\mathbf{v}_\theta(\mathbf{z}_t, t, c)$ conditioned on timestep $t$ and class $c$ to regress this velocity:
\begin{equation}\label{eq:vpred_loss}
\mathcal{L}_v = \mathbb{E}_{t, \mathbf{z}_0, \boldsymbol{\epsilon}} \| \mathbf{v}_\theta(\mathbf{z}_t, t, c) - \mathbf{v} \|^2.
\end{equation}
Generation integrates the learned ODE $\dot{\mathbf{z}}_t = \mathbf{v}_\theta$ from $t{=}0$ to $t{=}1$ via an Euler or Heun solver.

\subsection{$x$-Prediction on Standardized Features}\label{sec:xpred}

\noindent\textbf{Element-wise standardization.} DINOv2's per-token LayerNorm pins $\|\mathbf{z}\|^2 = d$ within each token, but leaves the \emph{cross-dataset} per-channel variance heterogeneous (${\gtrsim}10^2$ spread across channels, \S\ref{sec:experiments}). Before diffusion, we therefore standardize both patch tokens and the \texttt{[CLS]} token to zero mean and unit variance per element: $\tilde{z}_{c,h,w} = (z_{c,h,w} - \mu_{c,h,w}) / \sqrt{\sigma_{c,h,w}^2 + \epsilon}$ (analogously for $\mathbf{z}_\text{cls}$), using statistics precomputed on the training set. This diagonal preconditioner \cite{ahamed2025preconditioned} reduces the condition number $\kappa(\mathbf{H})$ of the data covariance and relaxes the near-constant-norm constraint that LayerNorm imposes on raw DINOv2 features \cite{kumar2025rjf}. The inverse transform is applied before decoding. Henceforth we use $\mathbf{z}$ to denote the standardized feature. We find this step is a \emph{prerequisite} rather than an optimization: training on raw DINOv2 features diverges entirely (Table~\ref{tab:ablation}).

\noindent\textbf{$x$-Prediction.} As established in \S\ref{sec:manifold}, DINOv2's norm concentration means linear flow-matching paths $\mathbf{z}_t = t\mathbf{z}_0 + (1{-}t)\boldsymbol{\epsilon}$ traverse ambient regions the encoder never outputs; at such off-manifold $\mathbf{z}_t$, the $v$-target acquires a radial component orthogonal to the data manifold---the \emph{geometric interference} phenomenon diagnosed by \citet{kumar2025rjf}, who resolve it with Riemannian Flow Matching \citep{chen2023flow} using SLERP paths. Under $v$-prediction, the network must fit this radial component and therefore spends capacity on the norm direction rather than the tangential (along-manifold) direction. We resolve the same problem more simply, by changing the output parameterization.

Setting $\mathbf{z}_0 = f_\text{enc}(\mathbf{x})$ to the standardized DINOv2 feature, \emph{$x$-prediction} \cite{li2025back} instead outputs $\hat{\mathbf{z}}_0 = f_\theta(\mathbf{z}_t, t, c)$ directly, with predicted velocity $\hat{\mathbf{v}}_\theta = (\hat{\mathbf{z}}_0 - \mathbf{z}_t)/(1-t)$; plugging this into the $v$-prediction loss (\ref{eq:vpred_loss}) yields
\begin{equation}\label{eq:loss_fm}
\mathcal{L}_\text{fm} = \mathbb{E}_{t, \mathbf{z}_0, \boldsymbol{\epsilon}} \| \hat{\mathbf{v}}_\theta - \mathbf{v} \|^2,
\end{equation}
\begin{wraptable}{r}{0.35\textwidth}
  \vspace{-15pt}
\centering
\small
\caption{\textbf{$x$-pred vs $v$-pred.} FID-50K, ImageNet $256^2$, Heun 50 steps, w/o guidance.}
\label{tab:xpred_vpred}
\setlength{\tabcolsep}{3pt}
\begin{tabular}{@{}lccc@{}}
\toprule
& 80 ep. & 200 ep. & 400 ep. \\
\midrule
$v$-pred & 3.17 & 2.11 & 1.86 \\
$x$-pred & \textbf{2.63} & \textbf{1.89} & \textbf{1.70} \\
\bottomrule
\end{tabular}
\vspace{-10pt}
\end{wraptable}
which is equivalent to the $x$-prediction loss $\|\hat{\mathbf{z}}_0 - \mathbf{z}_0\|^2$ up to a $(1{-}t)^{-2}$ reweighting (Appendix~\ref{app:xpred_equiv}). The $v$- and $x$-forms thus coincide as loss \emph{functionals}, but impose different learning problems on the network, because the output \emph{parameterization} determines which function is actually fit. Under $v$-prediction, the network must fit $(\mathbf{z}_0{-}\mathbf{z}_t)/(1{-}t)$: a target that depends on the off-manifold $\mathbf{z}_t$, diverges as $(1{-}t)^{-1}$ near $t{=}1$, and spans the full ambient space. Under $x$-prediction, the network fits $\mathbf{z}_0$: a target that lies on the low-dimensional data manifold by construction and does not depend on $t$ explicitly. The chord through off-manifold $\mathbf{z}_t$ persists at the network \emph{input}; at the \emph{output}, the target is confined to the data manifold rather than spanning the full ambient space. This manifold-targeting property is not itself DINOv2-specific \cite{li2025back}; what is unique here is the combination with DINOv2's isotropic per-coordinate variance and near-Gaussian marginals, which together bound the target and smooth its dependence on $\mathbf{z}_t$ (\S\ref{sec:manifold}), so a vanilla DiT suffices. Table~\ref{tab:xpred_vpred} confirms this empirically: with the same architecture, encoder, and noise schedule, $x$-prediction consistently outperforms $v$-prediction.

\subsection{Joint CLS--Patch Modeling}\label{sec:cls}

A unique advantage of operating in representation space is direct access to the \texttt{[CLS]} token---a global semantic summary that encodes category, layout, and appearance complementary to local patch content. In standard latent diffusion on VAE features such a global token is not part of the latent representation itself; in representation space, it is intrinsic. We model \texttt{[CLS]} jointly with patches in the same diffusion process: $\mathbf{z}_{\text{cls},t} = t\,\mathbf{z}_\text{cls} + (1 - t)\,\boldsymbol{\epsilon}_\text{cls}$ is projected, prepended to the patch sequence, and participates in bidirectional self-attention, aggregating spatial evidence into a global context and broadcasting refined guidance back to local tokens. A separate linear head produces the \texttt{[CLS]} prediction $\hat{\mathbf{z}}_{\text{cls},0}$, yielding an auxiliary $x$-prediction loss $\mathcal{L}_\text{cls} = \mathbb{E}\|\hat{\mathbf{v}}_{\text{cls},\theta} - \mathbf{v}_\text{cls}\|^2$ (written in velocity form for symmetry with Eq.~\ref{eq:vpred_loss}; equivalent to $\|\hat{\mathbf{z}}_{\text{cls},0}-\mathbf{z}_\text{cls}\|^2$ up to the same $(1{-}t)^{-2}$ reweighting of Appendix~\ref{app:xpred_equiv}) and total objective $\mathcal{L} = \mathcal{L}_\text{fm} + \lambda \mathcal{L}_\text{cls}$. During training, \texttt{[CLS]} noise is sampled independently from patch noise to avoid a $16^2{=}256\times$ variance collapse (\texttt{[CLS]} is a single vector while patch noise is $d{\times}16{\times}16$). At inference, we advance \texttt{[CLS]} and patches jointly with Heun + classifier-free guidance under (optionally separate) guidance scales; we use the same scale for both in all reported experiments, but the mechanism permits decoupling. Only patch tokens are decoded while \texttt{[CLS]} is discarded. We also couple the two noise streams at initialization via $\boldsymbol{\epsilon}_\text{cls} = \text{mean}_{h,w}(\boldsymbol{\epsilon})$ (\emph{coupled noise})---a minor but consistent improvement at convergence (\S\ref{sec:few_step}).

\subsection{Dimension-Aware Noise Schedule}\label{sec:noise}
\begin{wrapfigure}{r}{0.33\textwidth}
  \centering
  \vspace{-20pt}
  \includegraphics[width=0.33\textwidth]{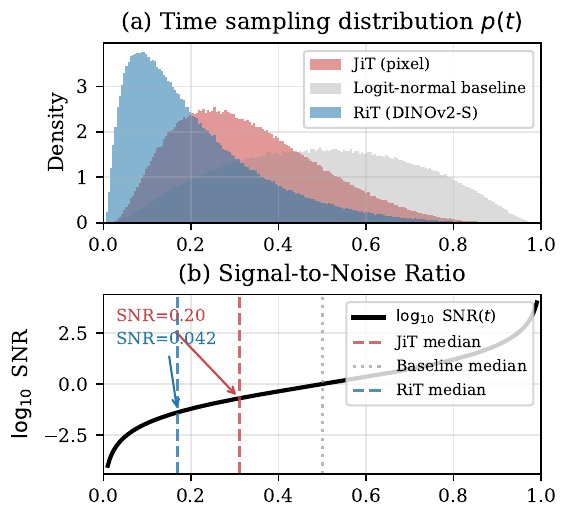}
  \captionsetup{font=footnotesize}
  \caption{\textbf{Time sampling $p(t)$ (top) and per-token SNR (bottom).}}
  \label{fig:noise_schedule}
  \vspace{-15pt}
\end{wrapfigure}
The SNR of $\mathbf{z}_t = t\mathbf{z}_0 + (1{-}t)\boldsymbol{\epsilon}$ is $\text{SNR}(t) = t^2/(1{-}t)^2$, but the \emph{effective} per-token SNR scales with the per-token dimension $d$ \citep{hoogeboom2023simple}: for a $d$-dimensional token, the $\ell_2$ noise magnitude grows as $\sqrt d$ while the signal stays at unit scale, so higher-$d$ tokens need lower $t$ (more noise) to reach the same relative corruption. A DINOv2-Small token has $d{=}384$, $128\times$ the per-pixel dimension of $3$, so a pixel-space schedule undertrains on noisy states. Following RAE \citep{zheng2025diffusion} and SD3 \citep{esser2024scaling}, we apply the dimension-dependent time shift $X' = Xs / (1 + (s{-}1)X)$ with $s = \sqrt{hwd / 4096} \approx 4.9$ to $X \sim \text{logit-}\mathcal{N}(0, 1)$ and set $t = 1 - X'$. This pushes the median $t$ from ${\approx}0.31$ to ${\approx}0.17$ ($5\times$ lower median SNR, Figure~\ref{fig:noise_schedule}); \S\ref{sec:experiments} shows this closes a $2\times$ FID gap over the pixel-space logit-normal baseline (3.17 $\to$ 1.44 at 800 epochs).

\section{Experiments}\label{sec:experiments}

\textbf{Setup.} RiT-XL has 28 layers, hidden dimension 1152, 16 attention heads, and FFN expansion ratio $4$, totaling 676M parameters. We train with a frozen DINOv2-Small encoder ($d{=}384$) and a pretrained RAE decoder \cite{zheng2025diffusion} on ImageNet, using 8 H200 GPUs ($\sim$12\,min per epoch). We evaluate with FID-50K using class-balanced sampling (50 images per class) following RAE \cite{zheng2025diffusion}. Full hyperparameters are in Appendix~\ref{app:architecture}.

\begin{figure}[!htb]
\centering
\begin{minipage}[t]{0.54\textwidth}
  \centering
  \captionsetup{font=footnotesize}
   \caption{\textbf{Convergence comparison on ImageNet $\mathbf{256^2}$.} FID-50K vs training epochs. }
  \includegraphics[width=\linewidth]{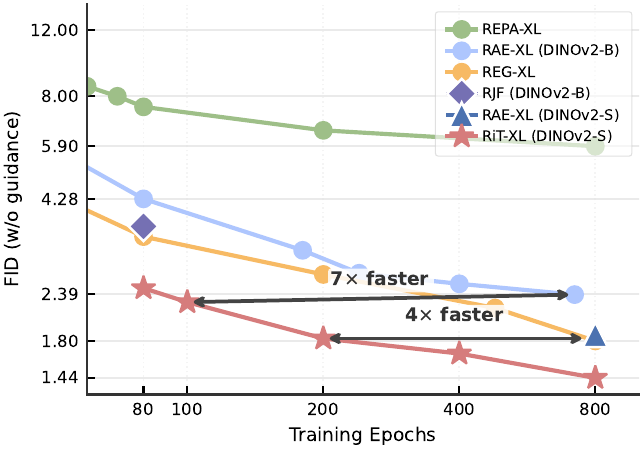}
 
  \label{fig:convergence}
\end{minipage}
\hfill
\begin{minipage}[t]{0.4\textwidth}
  \centering
  \footnotesize
  \captionsetup{font=footnotesize}
  \captionof{table}{\textbf{Ablation study.} Default: $x$-prediction (Table~\ref{tab:xpred_vpred}), element-wise standardization, time-shift schedule, $\lambda{=}0.2$, DINOv2-S. Each row replaces one factor. ``$\dagger$'' indicates training diverges.}
  \label{tab:ablation}
  \setlength{\tabcolsep}{3pt}
  \resizebox{\linewidth}{!}{
  \begin{tabular}{@{}lccc@{}}
  \toprule
  & 200 ep & 400 ep & 800 ep \\
  \midrule
  \rowcolor{black!8} Default (ours) & 1.83 & 1.67 & \textbf{1.44} \\
  \midrule
  \multicolumn{4}{@{}l}{\textit{Preconditioning}} \\
  \quad w/o standardization$^\dagger$ & 387.4 & 362.1 & 344.8 \\
  \multicolumn{4}{@{}l}{\textit{Noise schedule}} \\
  \quad Logit-normal & 4.45 & 4.11 & 3.17 \\
  \multicolumn{4}{@{}l}{\textit{CLS loss weight $\lambda$}} \\
  \quad w/o CLS & 1.89 & 1.70 & 1.63 \\
  \quad $\lambda{=}0.4$ & 1.86 & 1.64 & 1.50 \\
  \multicolumn{4}{@{}l}{\textit{Encoder}} \\
  \quad DINOv2-Base & 2.20 & 1.78 & 1.56 \\
  \bottomrule
  \end{tabular}
  }
\end{minipage}
\vspace{-15pt}
\end{figure}

\subsection{Convergence and Efficiency}

Figure~\ref{fig:convergence} compares RiT-XL against baselines. Against RAE-XL (DINOv2-S) \cite{zheng2025diffusion}---a $v$-prediction DiT-XL with the same encoder, decoder, and parameter count (676M) as RiT-XL, isolating the §\ref{sec:method} design choices---RiT-XL leads at every epoch and reaches FID 1.45 at 800 ep ($23\%$ better than 1.87). At 100 ep, RiT-XL already matches the larger RAE-XL (DINOv2-B) baseline at 720 ep ($7\times$ speedup); at 200 ep it matches RAE-XL (DINOv2-S) at 800 ep ($4\times$ speedup). RiT also surpasses the 800-ep FID of representation-alignment methods REPA \cite{yu2024representation} and REG \cite{wu2025representation} within 20--200 epochs. Concurrent RJF \cite{kumar2025rjf} tackles the same DINOv2 radial ambiguity via Riemannian Flow Matching on the norm-concentration sphere; at the matched 80-ep budget shown in Figure~\ref{fig:convergence}, RiT-XL reaches FID 2.48 (DINOv2-S) versus RJF's 3.62 (DINOv2-B).

\subsection{Every Recipe Choice Is Necessary}

Ablations use the full recipe unless noted, varying one factor at a time; we report ImageNet $256\!\times\!256$ FID-50K without guidance at Heun 50 steps.

\noindent\textbf{Element-wise standardization.} Raw DINOv2 features have heterogeneous per-channel variances (${\gtrsim}10^2$ range across channels). Training on raw features \emph{diverges}: the loss oscillates and FID stays at random-init level ($>300$) throughout training.

\noindent\textbf{Noise schedule.} The time-shift closes a $2\times$ FID gap over the original JiT logit-normal schedule (3.17 $\to$ 1.44 at 800 ep), confirming that reallocating training density toward higher noise is critical when per-token dimensionality grows (here $d{=}384$ vs.\ pixel's $d{=}3$).

\noindent\textbf{CLS token.} Without CLS modeling ($\lambda{=}0$), FID plateaus at 1.63; $\lambda{=}0.2$ reaches 1.44. Attention visualization (Appendix~\ref{app:cls_attention}) shows \texttt{[CLS]} aggregates coarse scene cues in early layers, integrates object--context relations in middle layers, and broadcasts refined guidance back in late layers.

\noindent\textbf{Encoder size.} Despite half the feature dimensionality, DINOv2-Small consistently beats DINOv2-Base (1.44 vs.\ 1.56 at 800 ep). Model capacity is not the bottleneck here: DINOv2-B has twice the ambient dimensionality at the same $\hat{d} \approx 33$, so the denoiser must regress over a higher-dimensional target without a corresponding gain in underlying structure. The Section~\ref{sec:manifold} analysis on DINOv2-Base is thus a \emph{conservative} characterization---the Small manifold used in main experiments is at least as favorable and the regression task is easier.

\subsection{Efficient ODE Convergence Enables Few-Step Generation}\label{sec:few_step}

The four geometric properties established in Section~\ref{sec:manifold}---high effective rank, well-conditioned covariance, near-Gaussian marginals, and on-manifold linear interpolants---jointly predict that the noise-to-data ODE should be \emph{efficiently solvable in few Heun steps} on DINOv2 features. We verify this directly by measuring Heun-solver truncation error in pixel space and show that it translates into order-of-magnitude gains under tight sampling budgets---without any distillation or consistency training.

\begin{table*}[htb]
\centering
% \vspace{-5pt}
\begin{minipage}[t]{0.75\textwidth}
\centering
\small
\captionsetup{font=footnotesize}
\caption{\textbf{Sampling schedule ablation.} FID-50K across six ODE time-discretization schedules and Heun step counts, with and without classifier-free guidance. Each cell reports independent\,/\,coupled noise FID. RiT-XL on DINOv2-S, 800 epochs.}
\label{tab:sample_schedule}
\renewcommand{\arraystretch}{1.05}
\setlength{\tabcolsep}{3pt}
\resizebox{\linewidth}{!}{
\begin{tabular}{@{}lcccc|cccc@{}}
\toprule
& \multicolumn{4}{c|}{\textit{w/o guidance} (CFG $= 1.0$, Heun steps)} & \multicolumn{4}{c}{\textit{w/ guidance} (CFG $= 3.7$, Heun steps)} \\
\textbf{Schedule} & 5 & 10 & 25 & 50 & 5 & 10 & 25 & 50 \\
\midrule
Uniform & 12.8\,/\,12.7 & 6.88\,/\,6.76 & 2.34\,/\,2.29 & 1.61\,/\,1.58 & 10.8\,/\,10.7 & 6.19\,/\,6.12 & 1.93\,/\,1.90 & 1.30\,/\,1.28 \\
EDM & \cellcolor{black!10}\textbf{2.37\,/\,2.34} & 1.61\,/\,1.58 & 1.49\,/\,1.47 & 1.47\,/\,1.46 & 2.01\,/\,1.99 & 1.33\,/\,1.32 & 1.17\,/\,1.15 & 1.16\,/\,1.14 \\
Cosine & 5.96\,/\,5.86 & 2.16\,/\,2.12 & 1.57\,/\,1.56 & 1.48\,/\,1.45 & 5.69\,/\,5.63 & 1.91\,/\,1.88 & 1.29\,/\,1.28 & 1.19\,/\,1.18 \\
Power-2 & 2.41\,/\,2.39 & 1.74\,/\,1.72 & 1.50\,/\,1.48 & 1.47\,/\,1.44 & \cellcolor{black!10}\textbf{1.99\,/\,1.98} & 1.48\,/\,1.46 & 1.18\,/\,1.16 & 1.16\,/\,1.15 \\
Log-SNR & 4.56\,/\,4.51 & 1.96\,/\,1.95 & 1.51\,/\,1.50 & 1.46\,/\,1.44 & 3.79\,/\,3.78 & 1.73\,/\,1.74 & 1.23\,/\,1.22 & 1.18\,/\,1.17 \\
Time-shift & 2.44\,/\,2.38 & \cellcolor{black!10}\textbf{1.59\,/\,1.58} & \cellcolor{black!10}\textbf{1.47\,/\,1.45} & \cellcolor{black!10}\textbf{1.46\,/\,1.44} & 1.99\,/\,1.99 & \cellcolor{black!10}\textbf{1.27\,/\,1.25} & \cellcolor{black!10}\textbf{1.15\,/\,1.14} & \cellcolor{black!10}\textbf{1.15\,/\,1.14} \\
\bottomrule
\end{tabular}
}
\end{minipage}
\hfill
\begin{minipage}[t]{0.24\textwidth}
\centering
\captionsetup{font=footnotesize}
\captionof{figure}{\textbf{Few-step FID at matched NFE}.}
\label{fig:nfe_compare}
% \vspace{-5pt}
\includegraphics[width=\linewidth]{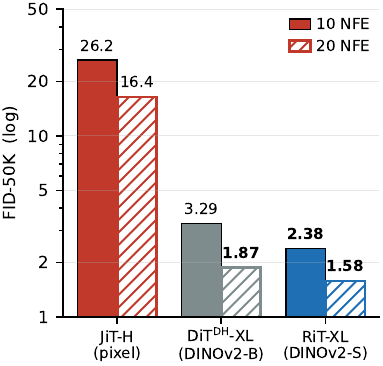}
\end{minipage}

\vspace{-10pt}
\end{table*}

\begin{wrapfigure}{r}{0.43\textwidth}
  \centering
  \vspace{-5pt}
  \includegraphics[width=0.43\textwidth]{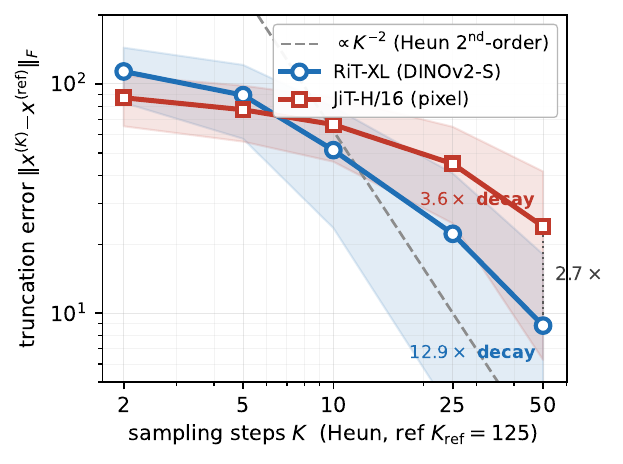}
  \captionsetup{font=footnotesize}
  \caption{\textbf{DINOv2 ODE converges in few Heun steps.} Pixel-space truncation error $\|x^{(K)}{-}x^{
          (\mathrm{ref})}\|_F$ vs step count $K$ (mean $\pm 1\sigma$ over 128 trajectories; each model vs its own $
          K_{\mathrm{ref}}{=}125$). RiT decays $12.9\times$ ($K{=}2{\to}50$) vs JiT's $3.6\times$; late-$K$ slope $
          -1.33$ vs $-0.91$, with the dashed line showing the Heun $\propto K^{-2}$ asymptote.}
  \label{fig:trajectory}
  \vspace{-15pt}
\end{wrapfigure}

\noindent\textbf{Pixel-space truncation error measurement.} For each model we generate 128 trajectories per space from matched $(\boldsymbol{\epsilon}, y)$ pairs, run Heun sampling at $K\in\{2,5,10,25,50\}$ plus a reference $K{=}125$, decode every endpoint to pixel space (RiT through the frozen RAE decoder), and measure the Frobenius distance $\|x^{(K)}{-}x^{(125)}\|_F$. Because each model is compared against its \emph{own} $125$-step reference, the metric isolates \emph{ODE convergence speed} from the absolute quality of either endpoint: a model that happens to converge to a poor fixed point is not artifactually rewarded. JiT uses the schedule reported in its own paper.

\noindent\textbf{RiT's truncation error decays $3.6\times$ steeper than JiT's.} Figure~\ref{fig:trajectory} shows the averages. RiT's truncation drops $12.9\times$ from $K{=}2$ to $K{=}50$ ($113.4{\to}8.8$), while JiT's drops only $3.6\times$ ($87.1{\to}23.9$). At $K{=}25$ (RiT's default), RiT is within $22.2$ Frobenius units of its $125$-step endpoint---consistent with Table~\ref{tab:sample_schedule}'s finding that RiT's $K{=}25$ FID of $1.47$ already matches the $K{=}50$ converged $1.46$. JiT at the same $K$ remains at $44.8$ Frobenius units and its FID is still far from converged (Figure~\ref{fig:nfe_compare}: JiT-H's $10$-NFE FID is $26.2$). Under the 2nd-order Heun error bound $\mathrm{err}(K)\!\propto\!\kappa_{\mathrm{eff}}(1/K)^2$, the $3.6{\times}$ gap in decay rate translates into a correspondingly smaller \emph{effective curvature} for DINOv2's marginal flow. This is the empirical counterpart of the four geometric properties quantified in Section~\ref{sec:manifold}: higher effective rank shortens the transport, Gaussianity matches the source distribution, tighter posterior lowers velocity-target variance, and on-manifold interpolants keep $\mathbf{z}_t$ in well-defined regions---each independently predicting a smoother, easier-to-integrate velocity field.

\noindent\textbf{Few-step generation.} Figure~\ref{fig:nfe_compare} isolates the role of representation space under matched NFE. At 10 NFE, pixel-space JiT-H yields FID 26.2 and DINOv2-space DiT$^\text{DH}$-XL yields 3.29, while RiT-XL reaches \textbf{2.38}---an order-of-magnitude improvement over pixel space and a clear gain over the DDT-equipped DINOv2 baseline; the same ordering holds at 20 NFE ($26.2{\to}16.4$ for JiT-H, $3.29{\to}1.87$ for DiT$^\text{DH}$-XL, $2.38{\to}\textbf{1.58}$ for RiT-XL). Within RiT itself (Table~\ref{tab:sample_schedule}), 5 Heun steps with a time-shift schedule reach FID 2.44 without guidance and 1.99 with guidance, 10 steps reach 1.59 and 1.25, and 25 steps already match full convergence---surpassing the majority of prior VAE-latent baselines in Table~\ref{tab:main_results}, which typically require ${\sim}250$ sampling steps.

\noindent\textbf{Sampling schedule ablation.} Table~\ref{tab:sample_schedule} compares six ODE time-discretization schedules (formal definitions in Appendix~\ref{app:sample_schedule}). At ${\geq}50$ steps, all non-uniform schedules converge to similar FID (1.43--1.45 w/o guidance, 1.14--1.15 w/ guidance), confirming the ODE is well-approximated. At 5 steps, the three concentrated schedules (EDM, power-2, time-shift: FID ${\approx}2.4$) outperform uniform spacing (12.7) by $5\times$, as they allocate more evaluations to the high-noise region where the velocity field varies most rapidly. Coupled noise (\S\ref{sec:cls}, independent\,/\,coupled cells in Table~\ref{tab:sample_schedule}) shifts FID by $<0.1$ uniformly across schedules and step counts; we adopt it in our main results for its small consistent gain but do not regard it as an essential component.

\noindent\textbf{From geometry to empirics.} The three mechanisms isolated in Section~\ref{sec:manifold} map directly onto RiT's gains. (i)~The $35\times$ better covariance conditioning ($\kappa{=}56$ vs.\ $2{,}000$) lets all variance directions train at comparable rates. (ii)~The tighter posterior $p(\mathbf{z}_0\mid\mathbf{z}_t)$ shrinks the irreducible target variance and contributes to the $7\times$ convergence speedup (Figure~\ref{fig:convergence}). (iii)~High effective rank (shorter transport), near-Gaussian marginals (smoother source$\to$data interpolation), and on-manifold interpolants (no void-crossing) yield low effective curvature, verified by the $3.6\times$ faster truncation-error decay (Figure~\ref{fig:trajectory}) and the few-step regime (Figure~\ref{fig:nfe_compare}). Pixel latents fail on all three and SD-VAE on at least two, so these mechanisms are empirically independent; their joint materialization on DINOv2 explains why a vanilla $676$M DiT surpasses $839$M DDT-equipped baselines both at convergence and under tight sampling budgets.

\begin{figure}[ht]
  % \vspace{-10pt}
  \centering
  \includegraphics[width=0.9\textwidth]{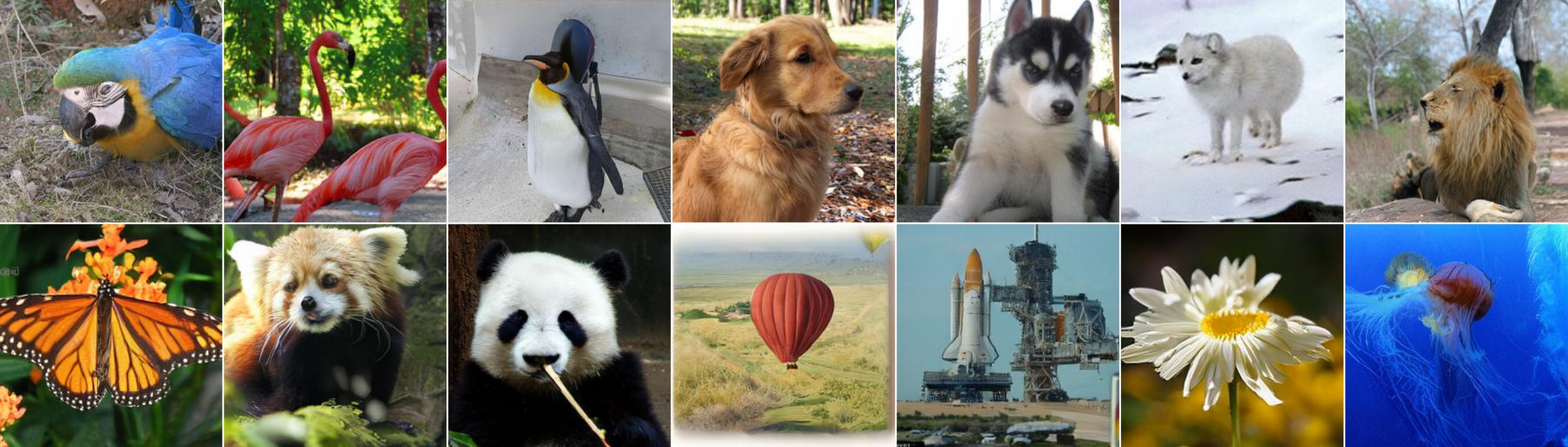}
  \caption{\textbf{Curated RiT-XL samples on ImageNet $256^2$} selected to span ImageNet categories.}
  \label{fig:samples_main}
  % \vspace{-10pt}
\end{figure}

\begin{table*}[ht]
\centering
\vspace{-7pt}
\small
\renewcommand{\arraystretch}{1.05}
\setlength{\tabcolsep}{4pt}
  % \captionsetup{font=footnotesize}
\caption{\textbf{Class-conditional image generation on ImageNet $\mathbf{256{\times}256}$.} With a vanilla DiT-XL backbone (676M parameters) and only 25 Heun steps, RiT achieves the best FID. Notably, RiT uses the smallest DINOv2 variant (DINOv2-S, $d{=}384$) among representation-based methods; DiT$^\text{DH}$-XL uses DINOv2-B, and FAE \cite{gao2025one} fine-tunes a DINOv2-G encoder, compressing its $d{=}1536$ features to a $d{=}32$ latent for generation.}
\label{tab:main_results}
\vspace{3pt}
\resizebox{\textwidth}{!}{
\begin{tabular}{l c c c cccc cccc}
\toprule
\multirow{2}{*}{\textbf{Method}} & \multirow{2}{*}{\textbf{AE}} & \multirow{2}{*}{\textbf{Epochs}} & \multirow{2}{*}{\textbf{\#Params}} & \multicolumn{4}{c}{\textbf{w/o guidance}} & \multicolumn{4}{c}{\textbf{w/ guidance}} \\
\cmidrule(lr){5-8} \cmidrule(lr){9-12}
 & & & & \textbf{FID}$\downarrow$ & \textbf{IS}$\uparrow$ & \textbf{Prec.}$\uparrow$ & \textbf{Rec.}$\uparrow$ & \textbf{FID}$\downarrow$ & \textbf{IS}$\uparrow$ & \textbf{Prec.}$\uparrow$ & \textbf{Rec.}$\uparrow$ \\
\midrule
\multicolumn{12}{l}{\textit{Pixel Diffusion}} \\
\arrayrulecolor{black!20}\midrule
ADM \citep{adm} & -- & 400 & 554M & 10.94 & 101.0 & 0.69 & 0.63 & 3.94 & 215.8 & \textbf{0.83} & 0.53 \\
PixelFlow-XL \citep{pixelflow} & -- & 320 & 677M & -- & -- & -- & -- & 1.98 & 282.1 & 0.81 & 0.60 \\
PixNerd-XL \citep{pixnerd} & -- & 320 & 700M & -- & -- & -- & -- & 1.93 & 298.0 & 0.80 & 0.60 \\
PixelDiT-XL \citep{pixeldit} & -- & 320 & 797M & -- & -- & -- & -- & 1.61 & 292.7 & 0.78 & 0.64 \\
JiT-G \citep{li2025back} & -- & 600 & 2B & -- & -- & -- & -- & 1.82 & 292.6 & 0.79 & 0.62 \\
\arrayrulecolor{black}\midrule
\multicolumn{12}{l}{\textit{Latent Diffusion}} \\
\arrayrulecolor{black!20}\midrule
DiT-XL \citep{peebles2023scalablediffusionmodelstransformers} & SD-VAE & 1400 & 675M & 9.62 & 121.5 & 0.67 & 0.67 & 2.27 & 278.2 & \textbf{0.83} & 0.57 \\
SiT-XL \citep{ma2024sit} & SD-VAE & 1400 & 675M & 8.61 & 131.7 & 0.68 & 0.67 & 2.06 & 270.3 & 0.82 & 0.59 \\
MaskDiT \citep{maskdit} & SD-VAE & 1600 & 675M & 5.69 & 177.9 & 0.74 & 0.60 & 2.28 & 276.6 & 0.80 & 0.61 \\
MDTv2-XL \citep{MDTv2} & SD-VAE & 1080 & 675M & -- & -- & -- & -- & 1.58 & \textbf{314.7} & 0.79 & 0.65 \\
REPA-XL \citep{yu2024representation} & SD-VAE & 800 & 675M & 5.78 & 158.3 & 0.70 & 0.67 & 1.29 & 306.3 & 0.79 & 0.64 \\
LightningDiT \citep{yao2025reconstruction} & VA-VAE & 800 & 675M & 2.17 & 205.6 & 0.77 & 0.65 & 1.35 & 295.3 & 0.79 & 0.65 \\
DDT-XL \citep{ddt} & SD-VAE & 400 & 675M & 6.27 & 154.7 & 0.68 & \textbf{0.69} & 1.26 & 310.6 & 0.79 & 0.65 \\
SVG-XL \citep{shi2025svg} & SVGTok & 1400 & 675M & 3.36 & 181.2 & -- & -- & 1.92 & 264.9 & -- & -- \\
REG-XL \citep{wu2025representation} & SD-VAE & 800 & 675M & 1.80 & - & - & -  & 1.36 & 299.4 & 0.77 & 0.66 \\
RAE-XL \citep{zheng2025diffusion} & RAE-DINOv2-S & 800 & 676M & 1.87 & 209.7 & 0.80 & 0.63 & 1.41 & 309.4 & 0.80 & 0.63 \\
% \rowcolor{black!8}
RAE$^\text{DH}$-XL\citep{zheng2025diffusion} & RAE-DINOv2-B & 800 & 839M & 1.51 & \textbf{242.9} & 0.79 & 0.63 & 1.16 & 261.0 & 0.77 & \textbf{0.67} \\
FAE-XL \citep{gao2025one} & FAE-DINOv2-G & 800 & 675M & 1.48 & 239.8 & \textbf{0.81} & 0.63 & 1.29 & 268.0 & 0.80 & 0.64 \\
% \rowcolor{black!8}
% {\color{gray}DiT$^\text{DH}$-XL$^{\ast}$} \citep{zheng2025diffusion} & {\color{gray}RAE-DINOv2-B} & {\color{gray}800} & {\color{gray}839M} & {\color{gray}1.51} & {\color{gray}242.9} & {\color{gray}0.79} & {\color{gray}0.63} & {\color{gray}1.13} & {\color{gray}262.6} & {\color{gray}0.78} & {\color{gray}0.67} \\
\arrayrulecolor{black}\midrule
\rowcolor{blue!5}
RiT-XL (ours) & RAE-DINOv2-S & 800 & 676M & \textbf{1.45} & 231.6 & \textbf{0.81} & 0.62 & \textbf{1.14} & 299.7 & 0.80 & 0.63 \\
\arrayrulecolor{black}\bottomrule
\end{tabular}
\vspace{-10pt}
}
\end{table*}

\subsection{Comparison with Prior Methods}

Table~\ref{tab:main_results} reports the full comparison. RiT's \emph{unguided} FID of 1.45 already surpasses the \emph{guided} FID of every method without a representation encoder (PixelDiT-XL 1.61, JiT-G 1.82, MDTv2-XL 1.58, DiT-XL 2.27, SiT-XL 2.06)---the well-structured semantic space captures the distribution so faithfully that CFG becomes far less necessary. Among representation-based methods, RiT achieves the best unguided FID (1.45 vs.\ DiT$^\text{DH}$-XL 1.51, FAE-DINOv2-G 1.48, RAE-XL 1.87) and the best guided FID (1.14 vs.\ DiT$^\text{DH}$-XL 1.28, FAE 1.29, REPA-XL 1.29). The margin over FAE is narrow at the unguided level (1.45 vs.\ 1.48), but obtained under a substantially simpler setup: FAE uses the largest DINOv2 variant (DINOv2-G, $d{=}1536$) and \emph{jointly fine-tunes} its encoder for generation, whereas RiT uses the smallest variant (DINOv2-S, $d{=}384$) with the encoder \emph{entirely frozen}---indicating that the geometric advantages of Section~\ref{sec:manifold} are already accessible off-the-shelf, without encoder co-adaptation. More fundamentally, FAE and RiT address \emph{different problems}: FAE eases generation by adapting the encoder to produce a more diffusion-friendly latent space (compressing DINOv2-G's $d{=}1536$ features to a $d{=}32$ generation latent), whereas RiT directly tackles modeling the existing high-dimensional representation distribution without altering the encoder. The two contributions are therefore largely orthogonal---encoder-side adaptation (FAE) and denoiser-side recipes (RiT) could in principle compose.

\noindent\textbf{RiT uses uniformly smaller components.} The denoiser is $676$M parameters ($19\%$ smaller than DiT$^\text{DH}$-XL's $839$M, no DDT head); the encoder is DINOv2-S ($d{=}384$, the \emph{smallest} DINOv2 variant), versus DiT$^\text{DH}$-XL's DINOv2-B ($d{=}768$) and FAE's DINOv2-G ($d{=}1536$). RiT also attains the highest unguided Precision ($0.81$) at competitive Recall ($0.62$), confirming that an off-the-shelf DINOv2 encoder and a vanilla backbone are sufficient when the representation distribution is favorable for flow matching.

\section{Related Work}

\noindent\textbf{Diffusion and flow matching for images.} Diffusion and flow matching \cite{ho2020denoising, song2020score, liu2022flow, esser2024scaling} underpin modern image generation. Latent diffusion \cite{rombach2022high} compresses images via a VAE; DiT/SiT \cite{peebles2023scalablediffusionmodelstransformers, ma2024sit} replace the U-Net with transformers. A critical design choice is the prediction target---$\epsilon$, $v$, or $x$. JiT \cite{li2025back} showed that $x$-prediction substantially outperforms the alternatives in pixel space by placing the target on the low-dimensional data manifold. Our work extends this insight by characterizing how a pretrained representation space embeds that manifold differently relative to $\mathcal{N}(\mathbf{0},\mathbf{I})$, and shows that DINOv2 is especially well-suited to $x$-prediction with a vanilla backbone.

\noindent\textbf{Leveraging representations for generation.}
VA-VAE, EQ-VAE, and Diffusability shape autoencoder training for diffusion-friendly latents \cite{yao2025reconstruction, kouzelis2025eq, skorokhodov2025improving}. REPA \cite{yu2024representation} adds a DINOv2 alignment loss that is active only during training. REG \cite{wu2025representation} addresses this train--inference gap by entangling a DINOv2 \texttt{[CLS]} into the SD-VAE trajectory; RiT differs in operating \emph{natively} in DINOv2 space where \texttt{[CLS]} is intrinsic. Neither REPA nor REG analyzes the manifold geometry, and both retain $v$-prediction on SD-VAE latents. RAE \cite{zheng2025diffusion} replaces the VAE with a DINOv2 encoder but adopts $v$-prediction and needs a DDT head for the ill-conditioned velocity field; concurrent RJF \cite{kumar2025rjf} instead uses Riemannian Flow Matching with SLERP paths on the norm-concentration sphere. We show $x$-prediction with element-wise standardization suffices to model DINOv2 features with a vanilla DiT---no architectural modification, no Riemannian reformulation.

\noindent\textbf{Few-step generation and distillation.} Progressive distillation \cite{salimans2022progressive}, consistency models \cite{song2023consistency, song2023improved}, and rectified flow \cite{liu2022flow} reduce sampling cost by training a dedicated few-step student or straightening the teacher's trajectories. These are orthogonal to RiT's contribution: RiT shows that the base model itself already reaches competitive few-step FID in a geometry-friendly representation space, without any distillation or consistency loss, and remains a natural teacher for such methods.

\noindent\textbf{Toward unified understanding--generation.} Unified vision models \cite{deng2025emerging, team2024chameleon, chen2025janus, tong2025metamorph, tong2026scaling} typically maintain \emph{separate} encoders for perception (CLIP/DINOv2) and synthesis (SD-VAE), with task-specific architectural components bolted onto the generative side. RiT's ability to generate competitively in DINOv2 space suggests a cleaner alternative: a single semantic representation and a single vanilla Transformer backbone---no DDT head, no Riemannian reformulation, no representation-alignment loss---can serve both tasks. The $7\times$ training speedup at matched encoder (\S\ref{sec:experiments}) and competitive few-step sampling (FID 2.0 at 5 Heun steps, 1.25 at 10 steps) further reduce the practical cost of attaching a generative head to an existing perception stack, making DINOv2-space RiT a promising base for unified pipelines in which the same features drive classification, retrieval, and synthesis.

\section{Conclusion}

We presented RiT, a vanilla DiT trained with $x$-prediction on frozen DINOv2 features that achieves FID 1.45 without guidance and 1.14 with classifier-free guidance on ImageNet $256^2$ using $19\%$ fewer denoiser parameters than DiT$^\text{DH}$-XL (676M vs.\ 839M) and the smallest DINOv2 variant (DINOv2-S, $d{=}384$); it further supports few-step generation (guided FID 2.0 at 5 Heun steps, 1.25 at 10 steps) without distillation or consistency training. The Section~\ref{sec:manifold} analysis indicates that representation-space diffusion becomes architecturally simpler whenever the feature distribution satisfies the four geometric axes we identify---high effective rank, well-conditioned covariance, near-Gaussian marginals, and on-manifold linear interpolants. Our results argue for a target-side reformulation ($x$-prediction) over architecture-side (DDT heads) or transport-side (Riemannian flow matching) solutions whenever the representation's distributional geometry is already favorable.

\section*{Acknowledgments}
We thank the Mila IDT team and their technical support for maintaining the Mila compute cluster. We also acknowledge the material support of NVIDIA in the form of computational resources. Throughout this project, Aishwarya Agrawal received support from the Canada CIFAR AI Chair award.

\bibliographystyle{plainnat}
\bibliography{references}

\clearpage
\appendix

\section{Limitations}\label{sec:limitations}

\noindent\textbf{DINOv2 encoder bias.} RiT inherits the inductive biases of the frozen DINOv2 encoder. DINOv2's SSL objective emphasizes semantic content over photometric detail, and prior work has observed weaker feature resolution on fine textures, thin structures, and small objects. These biases propagate directly into what RiT can generate, since the RAE decoder operates on the same features. Joint encoder fine-tuning (as in FAE \cite{gao2025one}) could mitigate this, at the cost of the simpler frozen-encoder setup we advocate.

\noindent\textbf{Class conditioning and resolution.} All RiT results are class-conditional on ImageNet at $256{\times}256$. We have not evaluated text-to-image generation, higher resolutions (e.g., $512$ or $1024$), or non-image modalities. The Section~\ref{sec:manifold} geometric analysis was measured on ImageNet images at DINOv2-Base and DINOv2-Small scales; whether the same four axes persist at larger model/data scales or under text conditioning is left to future work.

\noindent\textbf{Local-Gaussian assumption in the analysis.} The covariance-conditioning diagnostic (Section~\ref{sec:manifold}) rests on a local Gaussian approximation $p(\mathbf{z}_0) \approx \mathcal{N}(\boldsymbol{\mu}, \mathbf{H})$; the other three geometric axes are assumption-light but still report aggregate scalars that cannot rule out adversarial pockets of the manifold where the favorable properties fail. The flow-matching results empirically corroborate the geometric claims without establishing each axis as individually necessary for the observed efficiency gains.

\section{Equivalence of Velocity Loss and Reweighted $x$-Prediction Loss}\label{app:xpred_equiv}

We show that the velocity MSE loss with $x$-prediction parameterization (Eq.~\ref{eq:loss_fm}) is equivalent to a reweighted $x$-prediction loss. Given the forward process $\mathbf{z}_t = t\,\mathbf{z}_0 + (1-t)\,\boldsymbol{\epsilon}$, the target velocity is:
\begin{equation}
  \mathbf{v} = \mathbf{z}_0 - \boldsymbol{\epsilon} = \frac{\mathbf{z}_0 - \mathbf{z}_t}{1-t},
\end{equation}
where the second equality follows from substituting $\boldsymbol{\epsilon} = (\mathbf{z}_t - t\,\mathbf{z}_0)/(1-t)$. Under $x$-prediction, the network outputs $\hat{\mathbf{z}}_0 = f_\theta(\mathbf{z}_t, t, c)$, and the predicted velocity is $\hat{\mathbf{v}}_\theta = (\hat{\mathbf{z}}_0 - \mathbf{z}_t)/(1-t)$. Substituting both into the velocity loss:
\begin{align}
  \mathcal{L}_\text{fm}
  &= \mathbb{E}_{t,\mathbf{z}_0,\boldsymbol{\epsilon}} \left\| \hat{\mathbf{v}}_\theta - \mathbf{v} \right\|^2
  = \mathbb{E}_{t,\mathbf{z}_0,\boldsymbol{\epsilon}} \left\| \frac{\hat{\mathbf{z}}_0 - \mathbf{z}_t}{1-t} - \frac{\mathbf{z}_0 - \mathbf{z}_t}{1-t} \right\|^2
  = \mathbb{E}_{t,\mathbf{z}_0,\boldsymbol{\epsilon}} \frac{1}{(1-t)^2} \left\| \hat{\mathbf{z}}_0 - \mathbf{z}_0 \right\|^2.
\end{align}
Thus the velocity loss equals the $x$-prediction loss $\|\hat{\mathbf{z}}_0 - \mathbf{z}_0\|^2$ reweighted by $(1-t)^{-2}$, which upweights the loss at high $t$ (near clean data). The two losses are therefore equivalent as functionals; what differs is the network's \emph{parameterization}, which determines the function actually fit (\S\ref{sec:xpred}): $x$-prediction makes $\hat{\mathbf{z}}_0$ the direct network output, so its regression target always lies on the data manifold, whereas $v$-prediction asks the network to produce the ambient velocity $(\mathbf{z}_0 - \mathbf{z}_t)/(1-t)$, which depends on the off-manifold $\mathbf{z}_t$ and diverges as $t \to 1$.

\section{Manifold Analysis Details}\label{app:manifold}

This appendix provides formal definitions and implementation details for the manifold analysis metrics used in Section~\ref{sec:manifold}. All experiments are conducted on 10{,}000 randomly sampled ImageNet training images.

\paragraph{PCA spectrum and effective rank.}
We fit PCA with 512 components on the flattened feature vectors of each space. Let $\lambda_1 \geq \lambda_2 \geq \cdots \geq \lambda_k$ denote the eigenvalues of the sample covariance matrix and $\hat\lambda_i = \lambda_i / \sum_j \lambda_j$ the normalized eigenvalues. The \emph{effective rank} \citep{roy2007effective} is defined as
\begin{equation}
  \text{erank} = \exp\!\Bigl(-\sum_{i=1}^{k} \hat\lambda_i \log \hat\lambda_i\Bigr).
\end{equation}
It equals 1 when all variance concentrates in a single direction (maximally anisotropic) and equals $k$ when variance is perfectly uniform (maximally isotropic). For flow matching, higher effective rank means the data distribution is closer to the isotropic Gaussian source, requiring a less complex velocity field.

\paragraph{Intrinsic dimensionality (TwoNN).}
The TwoNN estimator \citep{facco2017estimating} estimates intrinsic dimensionality from the ratio of first- and second-nearest-neighbor distances. For each point $\mathbf{x}_i$, let $r_1^{(i)}$ and $r_2^{(i)}$ be the distances to its nearest and second-nearest neighbor, and $\mu_i = r_2^{(i)} / r_1^{(i)}$. Under the assumption that data is locally uniform on a $d$-dimensional manifold, the MLE estimator is
\begin{equation}
  \hat{d} = \frac{N}{\sum_{i=1}^{N} \log \mu_i},
\end{equation}
where $N$ is the number of valid samples with $\mu_i > 1$. We subsample 5{,}000 points and compute pairwise Euclidean distances in chunks to control memory.

\paragraph{Robustness of $\hat d$ at $D{\sim}10^5$.}
At such high ambient dimensionality, nearest-neighbor distances concentrate and any single-run intrinsic-dimension estimate can be noisy. We bootstrap TwoNN over 10 independent subsamples of 5{,}000 points, reporting the sample mean and standard deviation:
\begin{center}
\begin{tabular}{@{}lc@{}}
\toprule
Space  & $\hat d$ (mean $\pm$ std, 10 bootstraps) \\
\midrule
Pixel  & $33.6 \pm 1.3$ \\
DINOv2 & $32.6 \pm 0.8$ \\
\bottomrule
\end{tabular}
\end{center}
The pixel--DINOv2 gap of $1.0$ dimension is substantially below the combined standard deviation $\sqrt{1.3^2 + 0.8^2} \approx 1.5$ ($z\!\approx\!0.7$, $p\!\gg\!0.05$), so the two estimates are statistically indistinguishable. Larger-$k$ variants of the MLE estimator \citep{levina2004maximum} are known to suffer from upward bias at high ambient dimension \citep{facco2017estimating} and do not share this convergence property; we therefore rely on TwoNN as the primary estimator. DINOv2's advantage is not in manifold dimensionality but in the global geometry characterized by the other three axes of Section~\ref{sec:manifold}.

\paragraph{Marginal Gaussianity (excess kurtosis).}
For each dimension $j$, the excess kurtosis is
\begin{equation}
  \kappa_j = \frac{\mathbb{E}[(x_j - \mu_j)^4]}{\sigma_j^4} - 3,
\end{equation}
where $\mu_j$ and $\sigma_j$ are the per-dimension mean and standard deviation. A Gaussian distribution has $\kappa = 0$; positive values indicate heavier tails, negative values indicate lighter tails. We report the median of $|\kappa_j|$ across all dimensions as a scalar summary.

\paragraph{On-manifold interpolation score.}
For each intermediate frame along a linear interpolation path, we measure how well it stays on the natural image manifold via reconstruction error under a \emph{unified} pipeline: image $\to$ encode(DINOv2) $\to$ decode $\to$ MSE versus the input image. This pipeline is applied identically to both pixel-space and DINOv2-space interpolation frames:
\begin{itemize}[leftmargin=*,itemsep=1pt]
  \item \textbf{Pixel interpolation:} The intermediate frame $\mathbf{x}_t = (1{-}t)\mathbf{x}_a + t\,\mathbf{x}_b$ is passed through encode$\to$decode. Ghosting artifacts (off-manifold) produce high MSE because the encoder projects them to the nearest valid representation.
  \item \textbf{DINOv2 interpolation:} The intermediate representation $\mathbf{z}_t = (1{-}t)\mathbf{z}_a + t\,\mathbf{z}_b$ is first decoded to a pixel-space frame, which is then passed through the same encode$\to$decode pipeline.
\end{itemize}
By measuring both through the identical pipeline, the comparison is unbiased: the only difference is whether the frame was produced by pixel blending or DINOv2 latent interpolation. We average over 100 same-class pairs with 11 interpolation steps each.

\paragraph{Sanity check: encoder round-trip on DINOv2 interpolants.}
To rule out a trivial explanation that the DINOv2 interpolation pipeline enjoys near-zero reconstruction error by construction, we additionally measure whether $\mathbf{z}_t$ and $f_\text{enc}(f_\text{dec}(\mathbf{z}_t))$ are close in feature space. If the encoder merely re-projected arbitrary inputs to their nearest valid representation, the pixel-versus-DINOv2 MSE gap could be artifactually large. We report the average cosine similarity between $\mathbf{z}_t$ and its re-encoded counterpart, which remains high throughout the interpolation path; the gap in Figure~\ref{fig:manifold_analysis}(c) therefore reflects the off-manifold position of pixel blends rather than a baseline asymmetry in how the encoder treats each input.

\section{Architecture and Hyperparameters}\label{app:architecture}

\subsection{Model Architecture}

RiT uses a modernized DiT backbone (following JiT \cite{li2025back}/LightningDiT \cite{yao2025reconstruction}) operating on the $16{\times}16$ spatial grid of DINOv2 features. Our main experiments use DINOv2-Small ($d{=}384$); we also report DINOv2-Base ($d{=}768$) results in encoder ablations. Table~\ref{tab:arch} summarizes the model variants.

\begin{table}[h]
\centering
\caption{\textbf{RiT model variants.} All models use patch size 1, SwiGLU FFN (MLP ratio 4$\times$), QK-normalization, and VisionRoPE. Input dimension depends on the encoder: $384 \times 16 \times 16$ for DINOv2-Small (main experiments) or $768 \times 16 \times 16$ for DINOv2-Base.}

\label{tab:arch}
\small
\begin{tabular}{lcccccc}
\toprule
Model & Layers & Hidden dim & Heads & FFN dim & Params \\
\midrule
% RiT-S & 12 & 384 & 6 & 1536 & 33M \\
% RiT-B & 12 & 768 & 12 & 3072 & 130M \\
RiT-L & 24 & 1024 & 16 & 4096 & 458M \\
RiT-XL & 28 & 1152 & 16 & 4608 & 676M \\
\bottomrule
\end{tabular}
\end{table}

Each DiT block consists of:
\begin{enumerate}[leftmargin=*,itemsep=1pt]
    \item \textbf{adaLN modulation}: timestep and class embeddings are summed ($\mathbf{c} = \text{Emb}(t) + \text{Emb}(y)$) and projected to per-layer scale/shift parameters via a shared SiLU--Linear layer.
    \item \textbf{Multi-head self-attention} with QK-normalization (RMSNorm on Q and K before attention) and VisionRoPE for 2D spatial position encoding. \texttt{[CLS]} and register tokens are excluded from RoPE.
    \item \textbf{SwiGLU FFN}: $\text{FFN}(\mathbf{x}) = (\text{SiLU}(\mathbf{x} W_1) \odot \mathbf{x} W_3) W_2$.
\end{enumerate}
The final layer uses adaLN-modulated RMSNorm followed by a linear projection to $d$ output channels (384 for DINOv2-Small, 768 for DINOv2-Base). A separate linear head predicts the \texttt{[CLS]} token. The code supports attention and projection dropout applied only in the middle 50\% of layers; our main RiT-XL training sets both rates to $0$.

Following JiT \cite{li2025back}, we inject 32 learnable \emph{in-context tokens} at an intermediate layer (layer 8 for RiT-L). These tokens are initialized from the class embedding with added learnable positional embeddings, participate in self-attention for all subsequent layers, and are discarded before the final projection. They provide additional capacity for class-conditional generation without modifying the core DiT block.

\subsection{Training Hyperparameters}\label{app:train_hparams}

\begin{table}[h]
\centering
\caption{\textbf{Training hyperparameters} for RiT-XL on ImageNet $256^2$.}
\label{tab:train_hparams}
\small
\begin{tabular}{ll}
\toprule
Hyperparameter & Value \\
\midrule
Hardware & 8 $\times$ NVIDIA H200 GPUs \\
Throughput & $\sim$12 minutes per epoch \\
Optimizer & AdamW ($\beta_1{=}0.9$, $\beta_2{=}0.999$) \\
Base learning rate & $5 \times 10^{-5}$ (scaled by $\text{batch} / 256$) \\
LR schedule & Constant (after warmup) \\
Warmup epochs & 5 \\
Weight decay & 0.0 \\
Gradient clipping & $1.0$ (max $\ell_2$ norm) \\
Total epochs & 800 \\
Batch size & 1536 (8 $\times$ 192 per GPU) \\
EMA decay & 0.9999 / 0.9996 (dual tracking) \\
Label dropout & 0.1 \\
Attention / projection dropout & 0.0 (main configuration) \\
Noise schedule & Truncated logit-normal ($\mu{=}0$, $\sigma{=}1$) \\
Time shift $s$ & $\sqrt{16 \times 16 \times d / 4096}$ ($\approx 4.9$ for $d{=}384$) \\
Noise scale $\sigma_\epsilon$ & 1.0 \\
Epsilon clamp $\epsilon_t$ & 0.05 \\
CLS loss weight $\lambda$ & 0.2 \\
\bottomrule
\end{tabular}
\end{table}

\subsection{Sampling Hyperparameters}\label{app:sample_hparams}

\begin{table}[h]
\centering
\caption{\textbf{Sampling hyperparameters.}}
\label{tab:sample_hparams}
\small
\begin{tabular}{ll}
\toprule
Hyperparameter & Value \\
\midrule
ODE solver & Heun (2nd order) \\
Number of steps & 25 \\
CFG scale (patches) & 3.7 \\
CFG scale (CLS) & 3.7 \\
CFG interval & $[0.1, 0.98]$ \\
Epsilon clamp $\epsilon_\text{sample}$ & 0.05 \\
Generation precision & FP32 \\
EMA model & 0.9999 \\
\bottomrule
\end{tabular}
\end{table}

\subsection{Pseudocode}\label{app:pseudocode}

\begin{figure}[h]
\centering
\begin{minipage}[t]{0.48\textwidth}
\begin{lstlisting}[title={\small\textbf{Training}}]
# z0: data [B,C,H,W], z_cls: [B,C]
# dit: DiT model, y: class labels
t = sample_logit_normal(B, shift=s)
eps = randn_like(z0) * sigma
zt = t * z0 + (1 - t) * eps
# CLS noise: independent during training
eps_cls = randn_like(z_cls) * sigma
zt_cls = t * z_cls + (1 - t) * eps_cls
# x-prediction: predict clean data
z0_hat, cls_hat = dit(zt, t, y, zt_cls)
# convert to velocity for loss
v = (z0 - zt) / (1 - t).clamp_min(eps_t)
v_hat = (z0_hat - zt) / (1 - t).clamp_min(eps_t)
v_cls = (z_cls - zt_cls) / (1 - t).clamp_min(eps_t)
v_cls_hat = (cls_hat - zt_cls) / (1 - t).clamp_min(eps_t)
loss = mse(v_hat, v) + lam * mse(v_cls_hat, v_cls)
loss.backward()
\end{lstlisting}
\end{minipage}
\hfill
\begin{minipage}[t]{0.48\textwidth}
\begin{lstlisting}[title={\small\textbf{Sampling (Euler)}}]
# y: class labels, K: number of steps
z = randn(B, C, H, W) * sigma
z_cls = randn(B, C) * sigma
dt = 1.0 / K
for i in range(K):
    t = i / K
    # x-prediction -> velocity
    z0_hat, cls_hat = dit(z, t, y, z_cls)
    v = (z0_hat - z) / (1 - t).clamp_min(eps_t)
    v_cls = (cls_hat - z_cls) / (1 - t).clamp_min(eps_t)
    # classifier-free guidance
    # (omitted for clarity)
    z = z + dt * v
    z_cls = z_cls + dt * v_cls
# decode to image
x = rae_decoder(z)
\end{lstlisting}
\end{minipage}
\captionof{figure}{\textbf{PyTorch-style pseudocode} for RiT training (left) and sampling (right). The sampling code shows Euler for clarity; in all main experiments we use a 2nd-order Heun solver that additionally averages the velocity at $t$ and the predicted next step. Key differences from standard flow matching: $x$-prediction and joint CLS modeling.}
\label{fig:pseudocode}
\end{figure}

\section{Encoder Size Ablation}\label{app:encoder}

See Section~\ref{sec:manifold} (main text) for the encoder size ablation and Table~\ref{tab:ablation} for the full ablation results. DINOv2-Small ($d{=}384$) consistently outperforms DINOv2-Base ($d{=}768$) despite having half the feature dimensionality, reaching FID 1.44 vs.\ 1.56 at 800 epochs. The lower-dimensional latent space ($384 \times 16 \times 16$ vs.\ $768 \times 16 \times 16$) is easier for the denoiser to model, while DINOv2-Small still retains sufficient semantic information (TwoNN intrinsic dimensionality is comparable across encoder sizes).

\section{Uncurated Sample Grid}\label{app:samples}

\begin{figure}[h]
  \centering
  \includegraphics[width=\textwidth]{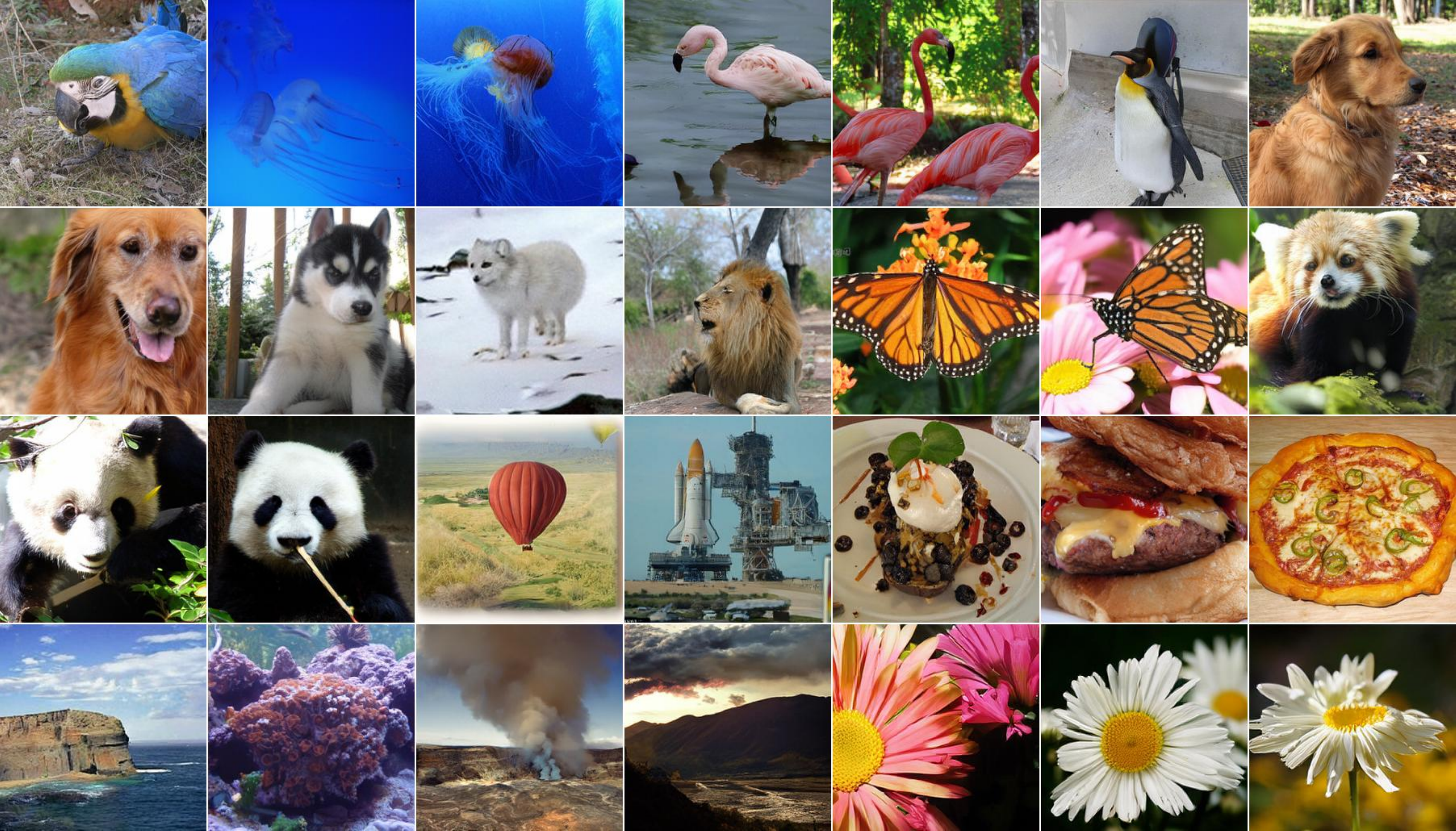}
  \caption{\textbf{Uncurated RiT-XL samples on ImageNet $256^2$.} 28 samples across diverse categories: macaw, jellyfish, flamingo, king penguin, golden retriever, Siberian husky, arctic fox, lion, monarch butterfly, red panda, giant panda, balloon, space shuttle, ice cream, cheeseburger, pizza, cliff, coral reef, volcano, and daisy. Generated with Heun 100 steps, CFG scale 3.7.}
  \label{fig:samples}
  \vspace{-3mm}
\end{figure}

Figure~\ref{fig:samples} shows uncurated samples generated by RiT-XL (DINOv2-Small encoder, 760 training epochs) using the Heun sampler with 100 steps and classifier-free guidance scale 3.7. The samples span diverse ImageNet categories including animals, food, landscapes, vehicles, and plants, demonstrating RiT's ability to produce high-fidelity, diverse images across a wide range of semantic categories.

\section{Sampling Schedule Analysis}\label{app:sample_schedule}

Table~\ref{tab:schedule_formulas} lists the six ODE time-discretization schedules evaluated in this work. Each schedule maps a normalized step index $i/K$ ($i=0,\dots,K$) to a timestep $t_i \in [0,1]$, where $t{=}0$ is pure noise and $t{=}1$ is clean data.

\begin{table}[h]
\centering
\small
\caption{\textbf{Sampling schedule definitions.} $K$ is the number of Heun steps and $i=0,\dots,K$.}
\label{tab:schedule_formulas}
\setlength{\tabcolsep}{6pt}
\begin{tabular}{@{}ll@{}}
\toprule
Schedule & Formula \\
\midrule
Uniform & $t_i = i/K$ \\[3pt]
Cosine & $t_i = \tfrac{1}{2}\bigl(1 - \cos(\pi\, i/K)\bigr)$ \\[3pt]
Log-SNR uniform & $t_i = \sigma\!\bigl(\text{logit}(\epsilon) + \tfrac{i}{K}[\text{logit}(1{-}\epsilon) - \text{logit}(\epsilon)]\bigr),\quad \epsilon{=}10^{-3}$ \\[3pt]
EDM \citep{karras2022elucidating} & $\sigma_i = \bigl(\sigma_{\max}^{1/\rho} + \tfrac{i}{K}(\sigma_{\min}^{1/\rho} - \sigma_{\max}^{1/\rho})\bigr)^\rho,\quad t_i = 1/(1{+}\sigma_i)$ \\
& $\sigma_{\min}{=}0.002,\; \sigma_{\max}{=}80,\; \rho{=}7$ \\[3pt]
Power-2 & $t_i = (i/K)^2$ \\[3pt]
Time-shift & $u_i = i/K,\quad t_i = 1 - u_i\, s\,/\,(1 + (s{-}1)\,u_i),\quad s{=}4.9$ \\
\bottomrule
\end{tabular}
\vspace{-3mm}
\end{table}

\begin{figure}[h]
  \centering
  \includegraphics[width=\textwidth]{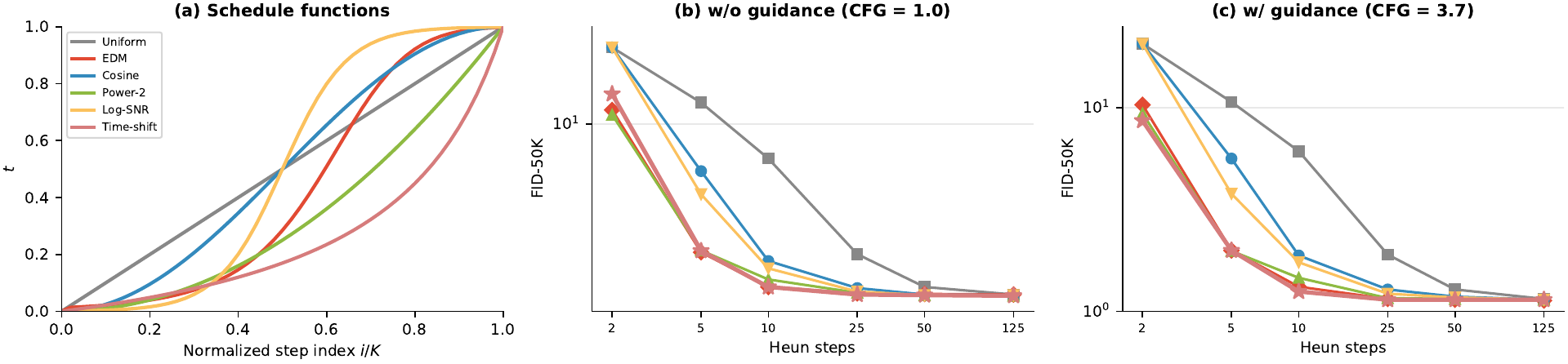}
  \vspace{-4mm}
  \caption{\textbf{Sampling schedule comparison.} (a)~Schedule functions $t(i/K)$ for $K{=}50$ steps. (b,\,c)~FID-50K vs.\ Heun step count without and with guidance. Coupled noise, RiT-XL on DINOv2-S, 800 epochs.}
  \label{fig:sample_schedule_curves}
  \vspace{-3mm}
\end{figure}

Figure~\ref{fig:sample_schedule_curves} visualizes these schedule functions and their effect on generation quality. Panel~(a) shows the schedule functions $t(i)$: uniform distributes steps evenly, while EDM, power-2, and time-shift concentrate steps near $t{=}0$ (the high-noise end); cosine and log-SNR are denser near both endpoints. Panels~(b) and~(c) show the corresponding FID as a function of Heun step count. Schedules that allocate more evaluations to the high-noise regime---where the velocity field varies most rapidly---achieve dramatically better FID at low step counts (${\leq}10$), while all non-uniform schedules converge at ${\geq}50$ steps.

\begin{table*}[h]
\centering
\small
\caption{\textbf{Full sampling schedule ablation (including 2-step).} FID-50K for ODE time-discretization schedules and Heun step counts. Each cell: independent\,/\,coupled noise. RiT-XL on DINOv2-S, 800 epochs.}
\label{tab:sample_schedule_ablation}
\renewcommand{\arraystretch}{1.05}
\setlength{\tabcolsep}{3pt}
\resizebox{\textwidth}{!}{
\begin{tabular}{@{}lcccccc|cccccc@{}}
\toprule
& \multicolumn{6}{c|}{\textit{w/o guidance} (CFG $= 1.0$, Heun steps)} & \multicolumn{6}{c}{\textit{w/ guidance} (CFG $= 3.7$, Heun steps)} \\
\cmidrule(lr){2-7} \cmidrule(l){8-13}
\textbf{Schedule} & 2 & 5 & 10 & 25 & 50 & 125 & 2 & 5 & 10 & 25 & 50 & 125 \\
\midrule
Uniform & 23.67\,/\,23.75 & 12.84\,/\,12.72 & 6.88\,/\,6.76 & 2.34\,/\,2.29 & 1.61\,/\,1.58 & 1.47\,/\,1.45 & 20.67\,/\,20.63 & 10.80\,/\,10.68 & 6.19\,/\,6.12 & 1.93\,/\,1.90 & 1.30\,/\,1.28 & 1.16\,/\,1.15 \\
EDM & 11.78\,/\,11.70 & \cellcolor{black!10}\textbf{2.37\,/\,2.34} & 1.61\,/\,1.58 & 1.49\,/\,1.47 & 1.47\,/\,1.46 & 1.47\,/\,1.45 & 10.43\,/\,10.34 & 2.01\,/\,1.99 & 1.33\,/\,1.32 & 1.17\,/\,1.15 & 1.16\,/\,1.14 & \cellcolor{black!10}1.14\,/\,\textbf{1.13} \\
Cosine & 23.67\,/\,23.75 & 5.96\,/\,5.86 & 2.16\,/\,2.12 & 1.57\,/\,1.56 & 1.48\,/\,1.45 & 1.45\,/\,1.44 & 20.67\,/\,20.63 & 5.69\,/\,5.63 & 1.91\,/\,1.88 & 1.29\,/\,1.28 & 1.19\,/\,1.18 & 1.15\,/\,1.14 \\
Power-2 & \cellcolor{black!10}\textbf{11.16\,/\,11.09} & 2.41\,/\,2.39 & 1.74\,/\,1.72 & 1.50\,/\,1.48 & 1.47\,/\,1.44 & 1.45\,/\,1.43 & 9.44\,/\,9.37 & \cellcolor{black!10}\textbf{1.99\,/\,1.98} & 1.48\,/\,1.46 & 1.18\,/\,1.16 & 1.16\,/\,1.15 & 1.15\,/\,1.14 \\
Log-SNR & 23.67\,/\,23.75 & 4.56\,/\,4.51 & 1.96\,/\,1.95 & 1.51\,/\,1.50 & 1.46\,/\,1.44 & 1.45\,/\,1.44 & 20.67\,/\,20.63 & 3.79\,/\,3.78 & 1.73\,/\,1.74 & 1.23\,/\,1.22 & 1.18\,/\,1.17 & 1.15\,/\,1.14 \\
Time-shift & 14.05\,/\,14.03 & 2.44\,/\,2.38 & \cellcolor{black!10}\textbf{1.59\,/\,1.58} & \cellcolor{black!10}\textbf{1.47\,/\,1.45} & \cellcolor{black!10}\textbf{1.46\,/\,1.44} & \cellcolor{black!10}\textbf{1.45\,/\,1.43} & \cellcolor{black!10}\textbf{8.59\,/\,8.64} & 1.99\,/\,1.99 & \cellcolor{black!10}\textbf{1.27\,/\,1.25} & \cellcolor{black!10}\textbf{1.15\,/\,1.14} & \cellcolor{black!10}\textbf{1.15\,/\,1.14} & 1.15\,/\,1.14 \\
\bottomrule
\end{tabular}
}
% \vspace{-3mm}
\end{table*}

\section{Random Samples}\label{app:random_samples}

\begin{figure}[!p]
  \centering
  \includegraphics[width=\textwidth]{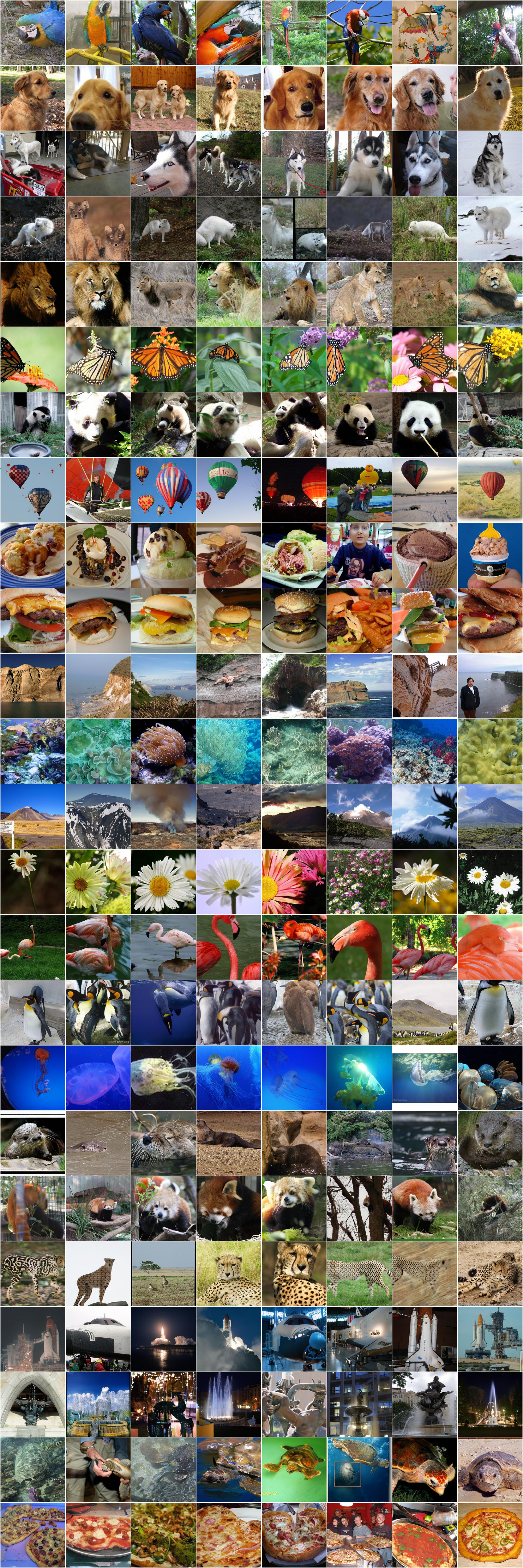}
  \caption{\textbf{Random (non-curated) RiT-XL samples on ImageNet $256^2$.} Each row shows 8 independently generated samples for a single class. 24 classes shown: macaw, golden retriever, Siberian husky, arctic fox, lion, monarch butterfly, giant panda, balloon, ice cream, cheeseburger, pizza, cliff, coral reef, volcano, daisy, flamingo, king penguin, jellyfish, otter, red panda, cheetah, space shuttle, fountain, and loggerhead turtle. Generated with Heun 100 steps, CFG scale 3.7.}
  \label{fig:random_samples}
\end{figure}

Figure~\ref{fig:random_samples} shows 192 randomly generated samples (8 per class, 24 classes) from RiT-XL without any curation or cherry-picking. Each row corresponds to a single ImageNet class. The model produces consistently high-quality and diverse samples across all categories.

\section{CLS--Patch Attention Analysis}\label{app:cls_attention}

\begin{figure}[!p]
  \centering
  \includegraphics[width=\textwidth]{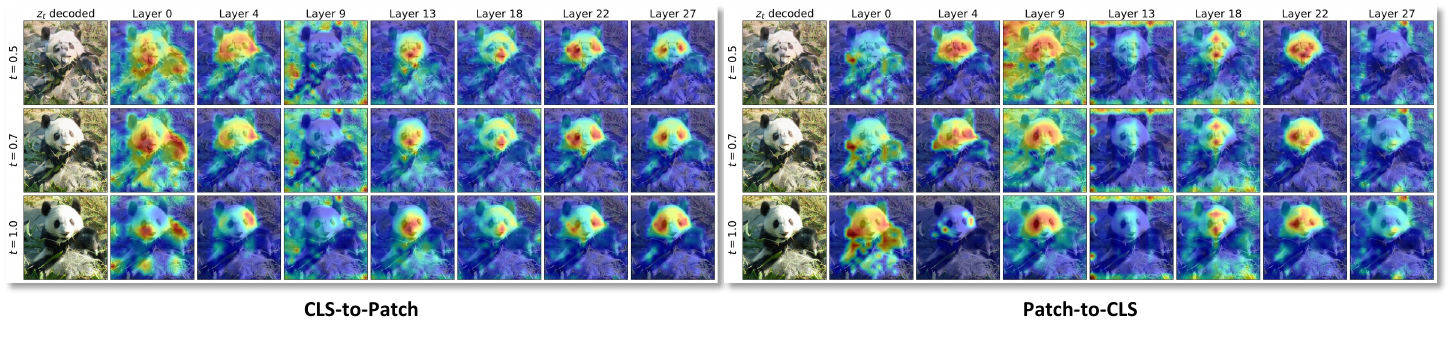}
  \caption{\textbf{Layer-wise CLS--patch communication in RiT.} Left: \texttt{[CLS]}-to-patch attention transitions from coarse scene aggregation to semantically salient regions. Right: patch-to-\texttt{[CLS]} attention shows global information exchange followed by focused refinement.}
  \label{fig:cls2patch_heat_map}
\end{figure}

Figure~\ref{fig:cls2patch_heat_map} visualizes the bidirectional attention between \texttt{[CLS]} and patch tokens across layers and timesteps. We observe a clear stage-wise communication pattern:

\textbf{CLS$\to$Patch (left).} In early layers, \texttt{[CLS]} attends broadly to salient foreground regions, aggregating coarse object cues. In middle layers, its attention expands to contextual/background regions, forming a global scene summary. In late layers, \texttt{[CLS]} re-focuses on semantically critical details (e.g., head and eyes), which strongly influence structural consistency and perceptual realism.

\textbf{Patch$\to$CLS (right).} Patch tokens increasingly query \texttt{[CLS]} at deeper layers, with the strongest reliance in semantically important regions, while low-information background patches rely less on it. These observations suggest that \texttt{[CLS]} acts as a global message hub: it collects distributed evidence, integrates object--context relations, and broadcasts refined global guidance back to patch tokens, improving object--background disentanglement and final generation quality.

\clearpage
% \newpage
% \input{checklist.tex}

\end{document}